\documentclass[preprint,10pt,3p]{elsarticle}

\usepackage{amssymb,mathrsfs}
\usepackage{graphicx,multirow,caption,subcaption,multicol,setspace,amsmath,wrapfig}
\usepackage[linesnumbered,ruled]{algorithm2e}
\usepackage{algorithmic}
\usepackage{adjustbox}
\usepackage{booktabs,caption}
\usepackage[flushleft]{threeparttable}
\usepackage{lineno}
\usepackage{enumitem}
\setlist{nosep}
\usepackage[table]{xcolor}
\usepackage{soul,xcolor}
\usepackage{amsfonts,tikz}
\usetikzlibrary{shapes}
\usepackage{makecell,eqnarray}
\newcommand{\xmark}{\ding{55}}%
\usepackage{amsthm}
\usepackage{bm}
\usepackage{wrapfig}
\theoremstyle{definition}

\usepackage{mathtools}

\graphicspath{ {Images/} }
\makeatletter
\newcommand{\algorithmfootnote}[2][\footnotesize]{%
  \let\old@algocf@finish\@algocf@finish
  \def\@algocf@finish{\old@algocf@finish
    \leavevmode\rlap{\begin{minipage}{\linewidth}
    #1#2
    \end{minipage}}%
  }%
}

\DeclareMathOperator*{\argmin}{arg\,min}

\definecolor{CC1}{rgb}{0.09, 0.45, 0.27}
\definecolor{CC2}{rgb}{0.9,0.45,0.1}
\usetikzlibrary{chains,decorations.pathreplacing}

\journal{European Journal of Operational Research}

\begin{document}
\begin{frontmatter}

\title{A Multi-criteria Approach to Evolve Sparse Neural Architectures\\ for Stock Market Forecasting}

\author[label1]{Faizal Hafiz\corref{cor1}}
\address[label1]{Artificial Intelligence Institute, SKEMA Business School, Université Côte d’Azur, Sophia Antipolis, France}
\ead{faizal.hafiz@skema.edu}
\cortext[cor1]{Corresponding author}

\author[label1]{Jan Broekaert}
\ead{jan.broekaert@skema.edu}

\author[label1]{Davide {La Torre}}
\ead{davide.latorre@skema.edu}

\author[label2]{Akshya Swain}
\address[label2]{Department of Electrical,Computer \& Software Engineering, The University of Auckland, New Zealand}
\ead{a.swain@auckland.ac.nz}

\begin{abstract}
This study
proposes a new framework to evolve \textit{efficacious} yet \textit{parsimonious} neural architectures for the movement prediction of stock market indices using technical indicators as inputs. In the light of a sparse signal-to-noise ratio under the Efficient Market hypothesis, developing machine learning methods to predict the movement of a financial market using technical indicators has shown to be a challenging problem. To this end, 
the neural architecture search is posed as a multi-criteria optimization problem to balance the efficacy with the complexity of architectures. In addition, the implications of different dominant trading tendencies which may be present in the pre-COVID and within-COVID time periods are investigated. An $\epsilon-$ constraint framework is proposed as a remedy to extract any concordant information underlying the possibly conflicting pre-COVID data. Further, a new search paradigm, Two-Dimensional Swarms (2DS) is proposed for the multi-criteria neural architecture search, which explicitly integrates sparsity as an additional search dimension in particle swarms. A detailed comparative evaluation of the proposed approach is carried out by considering genetic algorithm and several combinations of empirical neural design rules with a filter-based feature selection 
method (mRMR) as baseline approaches.
The results of this study convincingly demonstrate that the proposed approach can evolve parsimonious networks with better generalization capabilities.
\end{abstract}

\begin{keyword}
 Financial Forecasting \sep Neural Architecture Search \sep Multi-Criteria Decision Making \sep Feature Selection 
\end{keyword}

\end{frontmatter}

\section{Introduction}
\label{sec:introduction}


The quest for efficient forecasting algorithms for stock values and stock market indices has a long history, starting with the appraisal of stochastic fluctuations and evolving probabilities by~\citet{Regnault1863} and \citet{Bachelier1900}. The analysis of financial time series aims to correlate data points over time with a dependent output and, hence, provide a way to predict future values from historical data points. However, the quasi immediate information-adaptation mechanism underlying the \emph{Efficient Market Hypothesis} (EMH) severely reduces the signal-to-noise ratio in the financial time series \citep{Fama1965}, and, hence, caps from start the forecasting accuracy of any technical analysis algorithm. A broad consensus is, therefore, yet to be reached on the required technical formalisms as well as on the effectiveness of the existing forecasting algorithms despite a central-role of stock market trading in the make-up of financial products~\citep{Makridakis2018}. This study, in particular, focuses on the issues related to neural architectural design for a day ahead movement prediction of the NASDAQ composite index.

The relational information between samples in financial time series from stock trading prices or market composite indices remains precarious. The EMH questions the possibility to `beat the market', \textit{i.e.}, consistently predict the movement prior to the stock market's re-evaluation of the price. While the impossibility to hence capitalize on technical information to forecast future prices is supposed to be generally valid, recent studies have found occurrences of occasional or repetitive anomalies in this behavior, \textit{e.g.}, in the \emph{adaptive market hypothesis} approach~\citep{Urquhart2013} or the \emph{inefficient market hypothesis}~\citep{Asadi2012}. That said, any classifier  must learn from a mostly weak \emph{signal} within a dominant random progression of the stock market times series, which usually results in lower accuracy thresholds. 

Moreover, long-term anomalous events like the COVID-19 pandemic with global impact on the economies is to given extent reflected in market trading, resulting in distinct pre- and within-COVID trading tendencies~\citep{BuszkoEtAl2021,ChandraEtAl2021}. Taking into account the limited signal-to-noise ratio to be captured by technical indicators, the possible era-dependent disparate properties in the market time-series may hinder the \textit{learning} process, \textit{e.g}, the estimation of connection weights in neural networks. At the same time, it is not advisable to forfeit any concordant information contained in an otherwise possibly conflicting pre-COVID dataset, as the size of a training data set is decisive for the performance of a classifier. One of the main objectives of this study is, therefore, to investigate the implications of this scenario and possible remedies to reconcile the market behavior prior to and within-COVID time period into the learning process. 

Before we discuss the other objectives of this investigation, it is pertinent to briefly review some of the recent approaches to stock market prediction. Over the years, several methods and econometric techniques have been developed to estimate the future behavior of financial assets, which can be categorized into the following three main groups: classical time-series analysis~\citep{HyndmanEtAl2018}, technical indicators based models~\citep{Kim2003,Yu:Chen:2008,Huang2009,Kara2011,Asadi2012,Chang2012,Ticknor:2013,Nayak2015,kumar:Meghwani:2016,Zhong:Enke:2017,Lei2018,Tang2019,Dash2019,Bustos:2020,UlHaq2021}, and models with exogenous information~\citep{Wu2014,Xu2018,Zhong2019,Hoseinzade2019}. 

One of the proven classical tools for financial times series forecasting is the Auto Regressive Integrated Moving Average model (ARIMA), combined with exponential smoothing and de-trending and de-seasonalizing~\citep{HyndmanEtAl2018}. However, it has been argued that ARIMA analysis of financial data is often outperformed by Machine Learning (ML) approaches~\citep{Hewamalage2021}. 

Technical indicators have been the core data features to develop forecasting models and have given rise to many different ML based approaches of which we review more recent examples: \cite{Kim2003} compared SVM and ANN based prediction models using 12 technical indicators. \cite{Huang2009} implemented filter-based feature selection with a self-organizing feature map on 13 technical indicators resulting in a SVM-regressor to forecast the Taiwan index futures. In \citet{Kara2011}, ANN and SVM based prediction models were developed with 10 technical indicators to forecast the movement of the Istanbul stock exchange index. \cite{kumar:Meghwani:2016} proposed a model which uses a feature selection procedure and a proximal support vector machine (PSVM) to forecast one-day-ahead the direction of stock indices. In particular, four feature selection procedures were compared and contrasted to identify a subset of features from a pool of 55 technical indicators. In~\citet{Lei2018} a rough set based approach was used to select from a set of 15 indicators and to reduce number of hidden neurons in Wavelet Neural Networks (WNN), to forecast stock index movement. \cite{Dash2019} proposed a classifier ensemble using 6 technical indicators for data on three market indices.

Various studies have augmented the technical indicators with exogenous features derived from news media or economic sources~\citep{Wu2014,Xu2018,Zhong2019,Hoseinzade2019}. \citet{Zhong2019} tested deep neural networks to predict the daily return direction of stock market indices based on 60 financial and economic features. \citet{Wu2014} incorporated sentiment analysis to embed stock news articles as an auxiliary information. \citet{Xu2018} proposed the use of social media sentiment analysis using natural language processing. \citet{Hoseinzade2019} included various sources besides the market based indicators such as currency exchange rates, future contracts, commodity prices and other global market indices. 

It is worth emphasizing that, for the given data, the \textit{depth} of the neural architecture is an effective parameter of the neural architecture design. No preference for deep architecture should, therefore, be implemented a priori. The \textit{depth} and \textit{width} of neural architectures should be optimally linked to the task at hand~\citep{he2015deep}. More specifically, if an architecture is excessively deep for a given task, the initial layers will cover the optimal functional behavior and the final layers will attempt to simulate the \emph{identity} function hampered by the constraints set by its inherently non-linear activation components. Following these insights, the scope of this study is limited to \textit{shallow} neural architectures (as will be discussed in Section \ref{sec:NAS}).

While shallow neural networks have extensively been investigated for stock market prediction~\citep{Kim2003,Yu:Chen:2008,Kara2011,Ticknor:2013,Asadi2012,Zhong:Enke:2017}, the selection of an appropriate neural architecture is seldom addressed. This, in part, can explain often contradictory observations regarding the performance of neural networks, \textit{e.g.}, see~\citep{Kim2003,Kara2011}. In the context of neural networks, the selection of an architecture represents several topological decisions such as number of hidden layers and neurons as well as the selection of the activation function and the input features. It is easy to follow that such design choices are likely to have a significant impact on the network performance. In the context of stock prediction problem, often only the part of neural architecture search is addressed. For instance, \cite{Kara2011} and~\cite{Ticknor:2013} empirically tuned the number of hidden neurons whereas in the other investigation~\citep{Asadi2012,Zhong:Enke:2017} the selection of features is addressed, albeit \textit{independently} of the neural architecture. The other main objective of this study is, therefore, to identify an optimal neural architecture for stock market forecasting.

In particular, this study proposes an Extended Neural Architecture Search (ENAS), wherein the feature selection is viewed from the topological perspective as the selection of input neurons. This formulation allows for the simultaneous evolution of feature subsets along with the remaining neural architecture. The rationale here is to adjust the complexity of the downstream hidden layer architecture according to reduction in the input space which arises from the removal of redundant and/or noisy features. Further, ENAS is posed as a multi-objective problem wherein the goal is to evolve \textit{parsimonious} and \textit{efficacious} neural architectures. The motivation for this formulation is based on the \textit{principle-of-parsimony} or \textit{Occam's razor}~\citep{Rasmussen:Ghahramani:2001}, in a sense the goal is to identify a \textit{sparse} architecture from the pool of neural architectures with a similar generalization capability~\citep{Domingos:1999}. It is worth emphasizing that the search space of candidate neural architectures is intractable even for a moderate number of features and the other topological specifications. In particular, the feature selection is an NP-hard combinatorial problem since the feature correlation requires that each possible combination of features is examined~\citep{Guyon:Isabelle:2003,Blum:Langley:1997}. There is, thus, a clear need for an effective search approach to identify the optimal neural architectures.

This study proposes Two-Dimensional Swarms (2DS) to identify \textit{parsimonious} and \textit{efficacious} neural architectures. 2DS was originally developed by the authors to solve the feature selection problem in~\citep{Hafiz:Swain:2018}. In this study, we extend the key ideas of 2DS for the ENAS problem. In particular, the key attribute of 2DS is the direct and explicit integration of architecture complexity into the search process, \textit{i.e.}, instead of evolving selection likelihoods of only architecture aspects (such as number of hidden layers, neurons, features and activation functions), these are also evolved for distinct architectural complexities (discussed at length in Section~\ref{sec:search_opt_NN_top}). This key attribute is demonstrated to identify \textit{significantly} better neural architectures for the day-ahead prediction of the NASDAQ index direction of movement.

The search performance of 2DS is benchmarked by considering Genetic Algorithm as well as a combination of several distinct empirical neural design rules and a filter based feature selection algorithm, mRmR~\citep{Peng:Long:2005}, as the baseline approaches. Further, two learning scenarios are designed to test the hypothesis that disparate trading behavior -- reflected in a different NASDAQ index evolution -- prior to and within-COVID era may have \textit{adverse} effects on the weight estimation. An $\epsilon-$constraint framework is proposed to extract any useful information in the contradictory pre-COVID NASDAQ data while limiting any adverse effects on the network weight estimation for the within-COVID time window.

The rest of the article is organized as follows; Section \ref{sec:preliminaries} recalls the basics of forecasting financial time series and MCDM principles, the next Section \ref{sec:NAS} formulates the extended neural architecture search as well as provides the steps to determine the \textit{efficacy} and \textit{complexity} of candidate neural architectures. In Section \ref{sec:Nasdaq_COVID}, we discuss the possible effect of the COVID epidemic on NASDAQ trading and the ensuing multi-data set approach to avoid a type of inadvertent `data-poisoning' (or data subset incompatibility) and distinguish two main learning scenarios. With all criteria lined up, the implementation of the 2DS swarm search for the optimal network is put forward in Section \ref{sec:search_opt_NN_top}. The performance of the MCDM-2DS is evaluated and compared to benchmarks in Section \ref{sec:results}. In the conclusion Section \ref{sec:conclusions} we evaluate the learning scenarios and its generalization possibilities.

\section{Preliminaries}
\label{sec:preliminaries}

\begin{figure}[!t]
    \centering
        \begin{adjustbox}{max width=0.7\textwidth}
        \tikzstyle{block} = [rectangle, draw, fill=white, 
            text width=5em, text centered, rounded corners, minimum height=3em,line width=0.02cm]
        \tikzstyle{line} = [draw, -latex']
        \tikzstyle{sum} = [draw, fill=blue!20, circle, node distance=1cm]
        \tikzstyle{input} = [coordinate]
        \tikzstyle{output} = [coordinate]
        \tikzstyle{pinstyle} = [pin edge={to-,thick,black}]
        \begin{tikzpicture}[auto]
            \node [input, name=input] at (-1.3,0) {};
            \node [block, fill=orange!20] at (3,0) (TI) {Technical Indicators};
            \node [block, fill=blue!20] at (6.5,0) (NN) {Shallow Neural Network};
            \node [output] at (10,0) (op) {};
            \draw [->, thick] (input) -- node[align=center, below] {\{\textit{low, high, open}\\ \textit{close, volume}\}} node[align=center, above] {\textbf{Historical}\\ \textbf{Index Data}}(TI);
            
            \draw [->, thick] (TI.15) -- node[align=center, left] {}(NN.165);

            \draw [->, thick] (TI.345) -- node[align=center, left] {}(NN.195);
                        
            \draw [->, thick] (NN) -- node[align=center, above] {\textbf{Next Day}\\\textbf{movement}} node[align=center, below] {$\big\uparrow$ or $\big\downarrow$}(op);

        \end{tikzpicture} 
        \end{adjustbox}
        \caption{Day ahead prediction of index movement by technical analysis}
        \label{f:blockdia}
\end{figure}
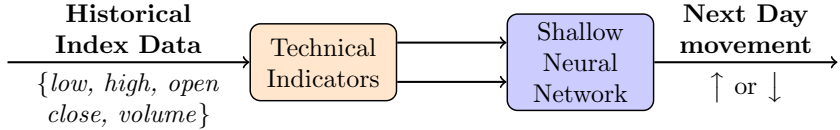 
\begin{table*}[!t]
  \centering
  \small
  \caption{Technical Indicators}
  \begin{adjustbox}{max width=\textwidth}
    \begin{tabular}{lll}
       \toprule
       \textbf{Financial Indicator} & \textbf{Parameter}  & \textbf{Expression}  \\[0.5ex]
       \midrule
       Opening price & -- &  $ O(t)$ \\[0.5ex]
       
       Highest intra-day price & -- &  $ H(t)$ \\[0.5ex]
       
       Lowest intra-day price & -- &  $ L(t)$ \\[0.5ex]
        
       Closing price & -- &  $ C(t)$ \\[0.5ex]

        Moving average & [5,10,15,20] &  $\mathit{MA}_n (t) = \frac{1}{n} \sum_{j=t-(n-1)}^t C(j)$ \\[0.5ex]
        
       Exponential moving average &  [5,10,15,20], $\alpha = \frac{2}{n+1} $ &  $\mathit{EMA}_n (t) =  \alpha  C(t) +(1-\alpha) \mathit{EMA}_{n} (t-1),  \ \ \   \mathit{EMA}_n (1)) = C(1)   $ \\[0.5ex]
        
        Relative strength index & [5,10,15,20] & $ \mathit{RSI}_n(t) =  \frac{\mathit{UPC}_n(t)/\mathit{UD}_n(t)}{\mathit{UPC}_n(t)/\mathit{UD}_n(t)+ \mathit{DPC}_n(t)/\mathit{DD}_n(t)} 100$\\[0.5ex]
        
        Stochastic index K & n=9 & $  K_n(t) = \frac{2}{3} K_n(t-1)+\frac{1}{3} \frac{C(t) -\mathit{HH}_n(t) }{\mathit{HH}_n (t) -\mathit{LL}_n (t)} $ \\[0.5ex]
        
        Stochastic index D & n=9 & $ D_n(t) = \frac{2}{3} D_n(t-1)+\frac{1}{3}K_n (t) $ \\[0.5ex]
        
        Moving average conv/div  & n=9 &  $ \mathit{MACD}_n(t) =  (1-\alpha) \mathit{MACD}_n(t-1)+\alpha\mathit{DIFF} (t) $\\[0.5ex]
        
        Larry Williams’ oscillator & [5,10,15,20] & $ \mathit{WR}_n(t) =\frac{\mathit{HH}_n(t) - C(t) }{\mathit{HH}_n (t) -\mathit{LL}_n (t)} 100 $\\[0.5ex]
        
        Psychological line & n=10 & $\mathit{PSY}_n(t) =  \frac{\mathit{UD}_n(t)}{n} 100$\\ [0.5ex]
        
        Price Oscillator  &   x = [5,10,15], y=[10,15,20]    &   $\mathit{OSCP}_{x,y}(t) = \frac{\mathit{MA}_x(t) - \mathit{MA}_y(t)}{\mathit{MA}_x(t)}$   \\ [0.5ex]
        
        Directional indicator up & [5,10,15,20]  & $\displaystyle \mathit{+DI}_n(t) =   \mathit{+DMS_{n}}(t)/ \mathit{TRS_{n}}(t) 100$\\ [0.5ex]
                                & \hfill where:  & $\mathit{+DMS_{n}}(t) =  \left( \mathit{+DMS_{n}}(t-1)\cdot (n-1)  +  ( H(t) - H(t-1) )  \right) /n  $\\ [0.5ex]
                                &               & $\mathit{TRS}_{n}(t) = \left( \mathit{TRS}_{n}(t-1)\cdot (n-1)  + \mathit{TR}(t)\right) / n$ \\[0.5ex]
                                &       &  $\mathit{TR}(t) =\max(H (t), C(t-1))-\min( L(t), C(t-1))$, \\ [0.5 ex] 

        Directional indicator down & [5,10,15,20]  & $\displaystyle \mathit{-DI}_n(t) = \mathit{-DMS_{n}}(t)/ \mathit{TRS_{n}}(t) \, 100 $ \\[0.5ex]
                                & \hfill where:  &  $\mathit{-DMS_{n}}(t)  =    ( \mathit{-DMS_{n}}(t-1)\cdot (n-1)  + ( L(t) -L(t-1))   ) /n $\\ [0.5ex]
        BIAS & n=5 &  $\mathit{BIAS}_n(t) = 100 \frac{C(t)-\mathit{MA}_n(t)}{\mathit{MA}_n(t)}$\\[0.5ex]
        
        Volume ratio & n=10 & $\mathit{VR}_n(t)  = \mathit{UV}_n(t)/(\mathit{UV}_n(t) + \mathit{DV}_n(t))$  \\[0.5ex]
        
        A ratio & n=20 & $  AR_n(t) =    \sum_{j=t-(n-1)}^t (H(j)-O(j))/ \sum_{j={t-(n-1)}}^t (O(j)-L(j))$  \\[0.5ex]
        
        B ratio  & n=20 & $  BR_n(t) =    \sum_{j=t-(n-1)}^t (H(j)-C(j))/ \sum_{j={t-(n-1)}}^t (C(j)-L(j))$   \\[0.5ex]
        Lowest Low          & n = 10 &  $\mathit{LL}_n(t)  = \min(L(t-n), \cdots, L(t-1) ) $ \\  [0.5ex]
        Highest High        & n = 10 &  $\mathit{HH}_n(t)  = \max(H(t-n), \cdots, H(t-1) ) $ \\  [0.5ex]
        Median Price        & n = 10 &  $\mathit{MP}_n(t)  = {\rm med}(C(t-n), \cdots, C(t-1) ) $ \\  [0.5ex]
        Average True Range  & n = 10 &  $\mathit{ATR}_n(t) =  (\mathit{ATR}_n(t)\cdot (n-1) + \mathit{TR}(t))/n$  \\  [0.5ex]
        Relative difference in percentage &  [5,10,15,20] & $\mathit{RDP}_n(t)= \frac{C(t)-C(t-n)}{C(t-n)} 100$  \\  [0.5ex]
        Momentum            &   [5,10,15,20] & $\mathit{MTM}_n(t) = C(t) - C(t-n)$  \\  [0.5ex]
        Price rate of change &   [5,10,15,20] & $\mathit{ROC}_n(t) = \frac{C(t) }{C(t-n)} 100 $  \\  [0.5ex]
        Ultimate Oscillator &   (x,y,z) = [10, 20, 30] & $\mathit{UO}_{x,y,z}(t) = \frac{100}{4+2+1}(4\, \mathit{AVG}(x)+ 2\, \mathit{AVG}(y) +   \mathit{AVG}(z) ) $  \\  [0.5ex]
                            & \hfill where: & $\mathit{AVG}(t) = \frac{\sum_{k=1}^t \mathit{BP}(k)}{\sum_{k=1}^t \mathit{TR}(k)}$, \\ [0.5 ex] 
                            &       &  $\mathit{BP}(t) = C(t) -\min( L(t), C(t-1))$ \\  [0.5ex]
        Ulcer Index     & n = 14 &  $\mathit{Ulcer}_n(t) = \sqrt{\sum_{k=1}^n R_k(t)^2/n } $ \\  [0.5ex]
                        & \hfill  where:  &   $ R_k(t) = \frac{100}{\mathit{HH}(t-k)} (C(t) - \mathit{HH}(t-k)) $ \\  [0.5ex]
        \bottomrule
    \end{tabular}%
 \end{adjustbox}
     \label{tab:indicators}
 \end{table*}
\subsection{Technical Indicators and ANN-based forecasting }
\label{subsec:technical_indicators}

Advanced data analysis techniques through ML have been applied to forecast financial time series. In practice mostly the two  approaches of technical and fundamental analysis are distinguished, relying on proper trading values (a chartist approach)  and financial and accounting factors respectively \citep{kirkpatrick2006,Oberlechner2001,NtiEtal2020}. While the combination of proper trading indicators and external time series data of, \textit{e.g.}, sentiment analysis, has been set forward, our model searches the optimal features within a space of technical indicators. 

Essentially, technical indicators are pre-defined functions of daily basic trading quantities of the opening, closing, daily high, daily low, prices and trading volume, which occur to be effective in forecasting trading prices or the upward or downward direction of a future trading price. Typically such indicators would identify trends or patterns which allow interpretation and inform trading decisions, in particular our model focuses on the daily direction of movement of the NASDAQ stock market, aiming to predict `$1$' for $C(t+1)-C(t)>0$ and `$0$' otherwise. In a general framework of ANN-based one-day-ahead prediction, the data samples are organized with a chosen number, $n_f$, of past-value technical indicators, $\mathit{TI}_{j}$, as features to predict next day's closing price; $\left(\mathit{TI} _{1}(t), \cdots, \mathit{TI} _{{n_f}}(t), C(t+1) \right)$, see Table \ref{tab:indicators}. Fig.~\ref{f:blockdia} gives the overall prediction framework being considered in this study.   

Technical indicators are construed to inform trading decisions based on diminishing noise volatility and capturing trends and patterns. Theoretically all technical indicators can be expressed for any time step $t$, in practice forecasting classifiers are developed and reported for 1-day time steps. Table \ref{tab:indicators} gives the technical indicators being considered in this study, which have been selected on the basis of previous investigation~\citep{Huang2009, kumar:Meghwani:2016}.

The indicators are derived from the following \textit{quantities} which are derived from daily trade values \citep{Huang2009, kumar:Meghwani:2016} : the number of upward days during the previous $n$ days, relative to the current day, $t$, is $\mathit{UD}_n(t)$,  similarly for downward days $\mathit{DD}_n(t)$; the cumulative closing values during  $n$ previous upward days relative to the current day $\mathit{UPC}_n(t)$, and similarly for the summation during downward days $\mathit{DPC}_n(t)$; the highest high price in the previous $n$ days, relative to current day $t$, is $\mathit{HH}_n(t)$, and similarly for the lowest low price  $\mathit{LL}_n(t)$; the exponential weighting factor is $\alpha = \frac{2}{n+1}$ for the period of $n$ previous days; the feature component $ \mathit{DIFF} (t) = \mathit{EMA}_{12} (t) - \mathit{EMA}_{26} (t) $  is the difference of the exponential moving average w.r.t. the current day $t$ for the periods $n=12$ and $n = 26$; the volume summation over $n$ previous days to current time step $t$ is $\mathit{TV}_n(t)$, while cumulative volume restricted to upward days is $UV_n(t)$ and for downward days $\mathit{DV}_n(t)$.

\subsection{Forecasting performance Measures} \label{subsec:performance_measures}

With a long-term secular drift on financial trading data,  a `\textit{brute} classifier', \textit{i.e.}, which deterministically always predicts this trend - could perform only marginally worse than a trained ANN when its classification performance is assessed by accuracy:
\begin{align}
{\rm Accuracy} & = \frac{TP + TN}{TP + FP + TN + FN}.
\end{align}
where, $TP$ and $TN$ respectively give the number of correctly predicted positive and negative class samples; and $FP$ and $FN$ indicates the cases otherwise.

Only upon inspection of the classification confusion matrix the inadequacy of the accuracy metric is apparent in lacking true negative (TN) cases. To remedy this erroneous performance interpretation, the Matthews correlation coefficient (MCC) has been proposed to give a more balanced  performance measure in binary classification~\citep{BoughorbelEtAl2017,Chicco2017,ChiccoEtAl2020}:
\begin{align}
{\rm MCC} &= \frac{TP \cdot TN - FP \cdot FN }{\sqrt{(TP+FP)(TP+FN)(TN+FN)(TN+FP)}} 
\end{align}
The MCC metric is more sensitive to all entries of the confusion matrix and essentially operates as  a correlation coefficient between the actual values and prediction outcomes. In our optimization approach the MCC is used to complement the accuracy objective to eliminate forecasting classifiers which converge to deterministic prediction, see Section \ref{sec:Nasdaq_COVID}.

\subsection{Multi-criteria Decision Making}
\label{subsec:MCDM}

Finding optimal solutions to decision making under multiple criteria with possibly conflicting tendencies, has been covered by MCDM approaches in Operations Research theory. A finite-dimensional MCDM problem with $n$ variables and  $p$ constraints reduces to the general solution statement: 
\begin{eqnarray}
\label{AP}
\min_{\mathcal{X} \in \Omega} \ J(\mathcal{X}).
\end{eqnarray}
where $\mathcal{X}$ is a compact subset of solutions $x$ in the parameter space $R^n$, and $J$ is a vector-valued map $J:\mathcal{X}\subset R^n\to R^p$.

We will apply the {\it linear scalarization} approach to reduce the MCDM configuration to a single criterion problem, by summing up all criteria with different weights which express the relative importance of each criterion:
\begin{eqnarray}
\label{eq:SP}
\min_{x\in\mathcal{X}} \sum_{i=1}^p \theta_i J_i(x),
\end{eqnarray}
where $\theta= (\theta_1, \ldots , \theta_p)$ is a vector taking values in $R^p_+$, the Pareto cone. 
Depending on the weight vector, solutions of the scalarized problem are also (weak) Pareto solutions of the multi-criteria problem, otherwise the found solutions are only partially satisfying  \citep{Sawaragi1985}. The present MCDM-2DS model is a performant approach adapted to the scalarized objective function,  a proper multi-criteria approach for forecasting time series with Pareto efficient solutions will be developed in a future study.

\section{Extended Neural Architecture Search: Simultaneous Evolution of Feature Subset and Neural Topology}
\label{sec:NAS}

To understand the neural architecture design problem, we begin by considering a pattern recognition problem over a dataset, $\mathcal{D}$, which has a total of $n_f$ input variables/features and $n_{class}$ output classes/labels, as follows:
\begin{small}
\begin{align}
    \label{eq:dataset}
    \mathcal{D} = \begin{Bmatrix} \left(x_1, x_2, \dots, x_{n_f}, y\right) \big| y \in (c_1,c_2,\dots, c_{n_{class}}) \end{Bmatrix}^{(k)}, \qquad k = 1,2,\dots \mathcal{N}
\end{align}
\end{small}
where, $\{x_1,x_2, \dots, x_{n_f}\} \in \mathcal{R}^{n_f}$; $c_i$ denotes the $i^{th}$ output class or label; and $\mathcal{N}$ denotes the total number of data patterns.

The first step of neural design for such problems is to estimate the `\textit{appropriate}' architectural complexity, \textit{i.e.}, \textit{how to determine the number of hidden neurons and hidden layers?} These issues stem from the fundamental design philosophy based on the \textit{principle-of-parsimony} wherein relatively less complex models are preferred from the given pool of models with similar generalization capabilities~\citep{Domingos:1999,Rasmussen:Ghahramani:2001}. The \textit{parsimonious} neural architectures have obvious advantages in terms of faster training, reduced requirements for computing and storage resources, and in many cases improved generalization capabilities over unseen data. 

In addition to the aforementioned \textit{topological} decisions, the other important issue is to select the non-linear processing capabilities of the hidden neurons, \textit{i.e.}, \textit{activation function}. The usual approach is to select a common activation function for the entire network, often referred to as a \textit{homogeneous network}. However,~\cite{Hagg:Mensing:2017} showed that by introducing another degree of freedom, in terms of selection of a distinct activation function for each hidden neuron, can lead to relatively parsimonious network architectures. Following these arguments, a candidate neural architecture (\textit{denoted by} $\mathcal{A}$) is defined as a set of $n_\ell-$\textit{tuples} in this study, as follows:
\begin{small}
\begin{align}
	\label{eq:nntuple}
  & \mathcal{A} \leftarrow \begin{Bmatrix} (s^1,f^1), (s^2,f^2), \dots, (s^{n_\ell}, f^{n_\ell}) & \Big | & s^k \in [0, s^{max}], & f^k \in \mathcal{F}, & \forall{k} \in [1,n_\ell] \end{Bmatrix}
\end{align}
\end{small}
where, $n_\ell$ denotes the \textit{maximum} number of hidden layers; $s^{max}$ gives the \textit{maximum} number of hidden neurons; $s^i$and $f^i$ denote the size (number of hidden neurons) and the activation function of the $i^{th}$ hidden layer; and $\mathcal{F}$ gives the set of possible activation functions. Accordingly, the search space of the possible candidate neural architectures ($ \Omega_\mathcal{A}$) is given by:
\begin{small}
\begin{align}
	\label{eq:omegaA}
  & \Omega_\mathcal{A}  = \Big( [0, s^{max}] \times \mathcal{F} \Big)^{n_\ell}
\end{align}
\end{small}
It is easy to follow that each hidden layer is represented by a corresponding tuple, $(s,f)$. Further, in this study, a distinct activation function ($f\in\mathcal{F}$) can be selected for each hidden layer, accordingly this formulation can evolve \textit{semi-heterogeneous} networks as per the categorization of~\cite{Hagg:Mensing:2017}.

Further, in this study we extend the scope of neural design to identify the subset of significant/relevant input features, $\{x_1, x_2,\dots, x_{n_f}\}$. Given that in practice many features are redundant and/or noisy, the identification of relevant features, also known as \textit{feature selection}, can lead to sparse models with a better generalization capability. To understand this further, consider a set of all features, $\rm X_{full}$:
\begin{small}
\begin{align}
    \label{eq:fs1}
    X_{\rm full} & = \begin{Bmatrix} x_1, & x_2, & \dots, & x_{n_f} \end{Bmatrix}
\end{align}
\end{small}
where, $x_i$ denotes the $i^{th}$ feature or the $i^{th}$ column of size $(\mathcal{N} \times 1)$ of the data-matrix $\mathcal{D}$. The objective of the feature selection is then to identify a \textit{sparse} feature subset, $\rm{X} \subset \rm{X}_{full}$, with equivalent or better classification performance. A detailed treatment on feature selection and different approaches to solve it can be found in~\citep{Blum:Langley:1997,Guyon:Isabelle:2003,Hafiz:Swain:2018}.

It is worth emphasizing that feature selection is a \textit{combinatorial problem}; the search for the optimal subset requires the examination of all possible feature combinations as the features are often correlated and cannot be evaluated independently~\citep{Blum:Langley:1997,Guyon:Isabelle:2003}. The search space of the feature selection problem ($\Omega_f$) is therefore given by,
\begin{small}
\begin{align}
    \label{eq:fs2}
    \Omega_F & = \Big\{ X \ | \ {X} \subset {X}_{\rm full} \wedge X \neq \emptyset \Big\} = 2^{{X}_{\rm full}} \setminus \{{X}_{\rm full}, \emptyset \} 
\end{align}
\end{small}
From the perspective of neural topology, the selection of relevant features can be viewed as the selection of input layer neurons, and therefore feature selection can be included within the framework of neural architecture design. However, most of the existing neural architecture search focuses only on the topological issues related to the design of hidden layers, for instance see~\citep{Stathakis:2009}. The simultaneous evolution of the \textit{feature subset} with the \textit{neural architecture} has the following direct benefits:

\begin{itemize}
    \smallskip
    \item \textit{From the neural topology perspective:} The simultaneous evolution removes the redundant and noisy inputs through feature selection. The consequent reduction in the input space allows for the exploration of relatively \textit{simpler} neural architectures, \textit{i.e.}, fewer hidden neurons and hidden layers.
    
    \smallskip
    
    \item \textit{From the feature selection perspective}: with the simultaneous evolution, the classification performance of the subsequent neural network can directly be used to evaluate a candidate feature subset. This approach, referred to as \textit{feature selection wrapper}~\citep{Guyon:Isabelle:2003,Hafiz:Swain:2018}, is often more precise than other indirect statistical or information theoretic based estimates of the classification performance (\textit{feature selection filters})~\citep{Guyon:Isabelle:2003}. A detailed discussion on \textit{filter} and \textit{wrapper} based feature selection can be found in~\cite{Guyon:Isabelle:2003,Hafiz:Swain:2018}.
    \smallskip
\end{itemize}

To embed the feature selection within the framework of neural architecture, this study proposes the Extended Neural Architecture Search (ENAS) which is formulated as a multi-objective optimization problem, as follows:
\begin{small}
\begin{align}
    \label{eq:ENS}
    \mathcal{X}^{\star} & = \argmin \limits_{ 
    \mathcal{X}_i \in \Omega}
    \begin{cases}
    \mathcal{E}(\mathcal{X}_i, \mathcal{D}_{test})\\
    \mathcal{C}(\mathcal{X}_i)
    \end{cases}\\
    \label{eq:comb2}
    \text{where, } \mathcal{X}_i & = \begin{Bmatrix} \mathcal{A}_i, & {X}_i \end{Bmatrix}, \quad \text{and} \quad \Omega = \Omega_A  \times \Omega_F
\end{align}
\end{small}
here $\mathcal{X}_i$ denotes the $i^{th}$ candidate solution which is a combination of neural architecture ($\mathcal{A}_i$) and feature subset (${X}_i$); $\Omega$ denotes the search space of the ENAS problem; $\mathcal{E}(\mathcal{X}_i, \mathcal{D}_{test})$ gives the \textit{classification error} of $\mathcal{X}_i$ over the test data $\mathcal{D}_{test}$; and $\mathcal{C}(\mathcal{X}_i)$ denotes the \textit{complexity} of $\mathcal{X}_i$. 

Following, the \textit{linear scalarization} approach (see Section~\ref{subsec:MCDM}), the multi-objective ENAS problem is transformed as follows:
\begin{small}
\begin{align}
    \label{eq:Jcwa}
    J(\mathcal{X}_i) & = \theta_\mathcal{E} \ \mathcal{E}(\mathcal{X}_i) \ + \theta_\mathcal{C} \ \mathcal{C}(\mathcal{X}_i)
\end{align}
\end{small}
where, $[\theta_\mathcal{E}, \ \theta_\mathcal{C}]$ denote the \textit{preference weights} specified by the Decision Maker (DM). The following three preference scenarios are being considered:
\begin{enumerate}
    \item Efficacy over Complexity: $\Theta_1 =[\theta_\mathcal{E}, \ \theta_\mathcal{C}] = [0.75, \ 0.25]$
    \item Balanced Scenario: $\Theta_2 =[\theta_\mathcal{E}, \ \theta_\mathcal{C}] = [0.50, \ 0.50]$
    \item Complexity over Efficacy: $\Theta_3 =[\theta_\mathcal{E}, \ \theta_\mathcal{C}] = [0.25, \ 0.75]$
\end{enumerate}
These preference weights were determined using the multiplicative preference relations approach in~\citep{Zhang:Chen:2004,Hafiz:Swain:MOEA:2020}.

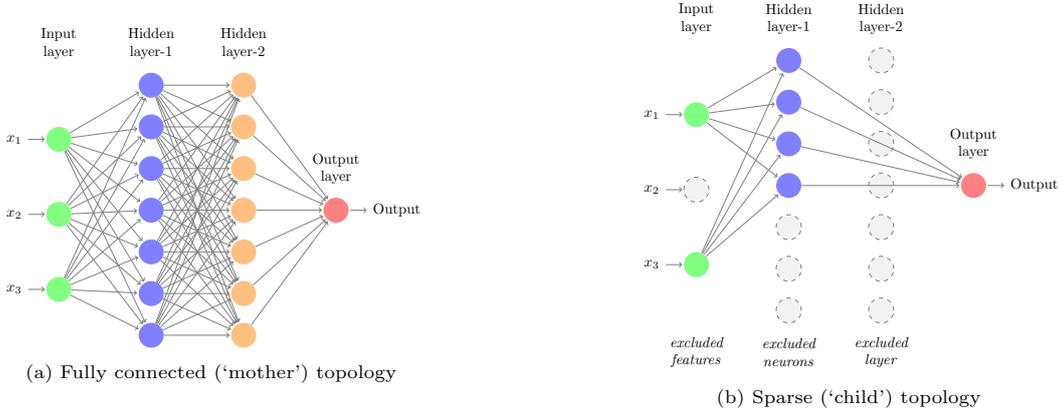
\begin{figure}[!t]
\centering

\begin{subfigure}{.49\textwidth}
   
   \centering
    \begin{adjustbox}{max width=0.7\textwidth}
    \def\layersep{2.2cm}

    \begin{tikzpicture}[shorten >=1pt,->,draw=black!50, node distance=\layersep]
        \tikzstyle{every pin edge}=[<-,shorten <=1pt]
        \tikzstyle{neuron}=[circle,fill=black!25,minimum size=17pt,inner sep=0pt]
        
        
        \tikzstyle{input neuron}=[neuron, fill=green!50];
        \tikzstyle{output neuron}=[neuron, fill=red!50];
        \tikzstyle{hidden neuron}=[neuron, fill=blue!50];
        \tikzstyle{hidden neuron2}=[neuron, fill=orange!50];
        \tikzstyle{annot} = [text width=4em, text centered]
    
        \foreach \name / \y in {1,...,3}
            \node[input neuron, pin=left:$x_\y$] (I-\name) at (0,-1.8*\y) {};
    
        \foreach \name / \y in {1,...,7}
            \path[yshift=0.5cm]
                node[hidden neuron] (H-\name) at (\layersep,-\y cm) {};
        
        \foreach \name / \y in {1,...,7}
            \path[yshift=0.5cm]
                node[hidden neuron2] (H2-\name) at (2*\layersep,-\y cm) {};
    
        \node[output neuron,pin={[pin edge={->}]right:Output}, right of=H2-4] (O) {};
    
        \foreach \source in {1,...,3}
            \foreach \dest in {1,...,7}
                \path (I-\source) edge (H-\dest);
        
         \foreach \source in {1,...,7}
            \foreach \dest in {1,...,7}
                \path (H-\source) edge (H2-\dest);
    
        \foreach \source in {1,...,7}
            \path (H2-\source) edge (O);
    
        \node[annot,above of=H-1, node distance=1cm] (hl) {Hidden layer-1};
        \node[annot,left of=hl] {Input layer};
        \node[annot,right of=hl] {Hidden layer-2};
        \node[annot,above of=O, node distance=1cm] {Output layer};
    \end{tikzpicture}
    \end{adjustbox}
    \caption{Fully connected (`mother') topology}
    \label{f:nna1}
   
\end{subfigure}
\hfill
\begin{subfigure}{.49\textwidth}
    \centering
    \begin{adjustbox}{max width=0.7\textwidth}
    \def\layersep{2.2cm}

    \begin{tikzpicture}[shorten >=1pt,->,draw=black!50, node distance=\layersep]
        \tikzstyle{every pin edge}=[<-,shorten <=1pt]
        \tikzstyle{neuron}=[circle,fill=black!25,minimum size=17pt,inner sep=0pt]
        \tikzstyle{input neuron}=[neuron, fill=green!50];
        \tikzstyle{output neuron}=[neuron, fill=red!50];
        \tikzstyle{hidden neuron}=[neuron, fill=blue!50];
        \tikzstyle{missing neuron}=[neuron, fill=gray!10,draw,dashed];
        \tikzstyle{annot} = [text width=4em, text centered]
    
        \foreach \name / \y in {1,3}
            \node[input neuron, pin=left:$x_\y$] (I-\name) at (0,-1.8*\y) {};
            
        \foreach \name / \y in {2}
            \node[missing neuron, pin=left:$x_\y$] (I-\name) at (0,-1.8*\y) {};
            
        \foreach \name / \y in {1,...,4}
            \path[yshift=0.5cm]
                node[hidden neuron] (H-\name) at (\layersep,-\y cm) {};
        
        \foreach \name / \y in {5,...,7}
            \path[yshift=0.5cm]
                node[missing neuron] (H-\name) at (\layersep,-\y cm) {};
        
        \foreach \name / \y in {1,...,7}
            \path[yshift=0.5cm]
                node[missing neuron] (H2-\name) at (2*\layersep,-\y cm) {};
    
        \node[output neuron,pin={[pin edge={->}]right:Output}, right of=H2-4] (O) {};
    
        \foreach \source in {1,3}
            \foreach \dest in {1,2,3,4}
                \path (I-\source) edge (H-\dest);
    
        \foreach \source in {1,2,3,4}
            \path (H-\source) edge (O);
    
        \node[annot,above of=H-1, node distance=1cm] (hl) {Hidden layer-1};
        \node[annot,left of=hl] {Input layer};
        \node[annot,right of=hl] {Hidden layer-2};
        \node[annot,above of=O, node distance=1cm] {Output layer};
        \node[annot,below of=H2-7, node distance=1cm] (bhl2) {\textit{excluded layer}};
        \node[annot,left of=bhl2] (bhl1) {\textit{excluded neurons}};
        \node[annot,left of=bhl1] {\textit{excluded features}};
    \end{tikzpicture}
    \end{adjustbox}
    \caption{Sparse (`child') topology}
    \label{f:nna2}
\end{subfigure}

\caption{Fully connected vs candidate sparse neural topologies. Grey and dotted circle represent excluded/skipped features or neurons.}
\label{f:nnarch}
\end{figure}
\subsection{Estimating Complexity of Neural Architectures \label{subsec:NeuComp}}

It is clear, the goal of the ENAS problem is to identify an \textit{efficacious} and \textit{parsimonious} combination of neural architecture and feature subset. To this end, we begin by formulating a function to quantify neural \textit{complexity} as follows:
\begin{small}
\begin{align}
   \label{eq:complexity2}
   \mathcal{C}(\mathcal{X}_i) & = \frac{1}{3} \left\{ \frac{\# (X_i)}{n_f} + \frac{\# \Big( \begin{Bmatrix} (s^k,f^k) \ \big| \ s^k \neq 0, & \forall k \in [1,n_\ell] \end{Bmatrix} \Big)}{n_\ell} + \sum \limits_{k=1}^{n_\ell} \frac{s^k}{s^{max}} \right\}
\end{align}
\end{small}
where, $\mathcal{X}_i$ denotes the candidate solution under consideration; $X_i$ gives the feature subset embedded within $\mathcal{X}_i$; $s^{max}$ and $n_\ell$ respectively give the maximum number of hidden neurons and hidden layers; $n_f$ denotes the total number of features; and $\#(\cdot)$ determines the \textit{cardinality} of a particular set. 

It is clear that the neural complexity, $\mathcal{C}(\cdot)$, is formulated as a three parts function of various topological attributes to account for number of features, hidden layers and hidden neurons encoded into a candidate solution. A lower value of $\mathcal{C}$ is desirable as it implies a reduced \textit{complexity} or increased \textit{parsimony}. The complexity is bounded, $\mathcal{C} \in [0,1]$, and it will achieve its maximum value of `$1$' for the \textit{mother}-topology which includes the full feature set as well as the maximum allowable hidden layers and hidden neurons.

To understand this further, consider a \textit{naive} pattern recognition problem with the following specifications: \textit{number of features}, $n_f = 3$; \textit{maximum hidden neurons}, $s^{max} = 7$; and \textit{maximum number of hidden layers}, $n_\ell = 2$. The fully-connected \textit{mother} topology ($\mathcal{C} = 1$) for these specification is shown in Fig.~\ref{f:nna1}. In contrast, Fig.~\ref{f:nna2} gives a possible \textit{sparse} candidate topology wherein a reduced feature subset, $X = \{x_1, x_3\}$, is selected along with a relatively fewer neurons in the first hidden layer ($s^1=4$) and the second hidden layer is \textit{skipped} (\textit{i.e.}, $s^2 = 0$). Accordingly, following~(\ref{eq:complexity}), the complexity of the sparse topology is determined to be $\mathcal{C} = 0.58$.

\subsection{Estimating Efficacy of Neural Architectures \label{subsec:NeuEff}}

Next, we focus on estimating the \textit{efficacy} of the candidate solution. We begin by noting that the neural architecture search is approached here as a \textit{bi-level} optimization problem, \textit{i.e.}, while the search for neural architecture continues at a higher level, a \textit{gradient-descent} algorithm estimates the corresponding network connection weights ($\mathcal{W}$) at the lower level. The estimation of $\mathcal{W}$ is often considered a \textit{noisy} function,  since it is influenced by various factors including but not limited to network initialization and the capability of the gradient-descent algorithm to avoid local minima. To account for these limitations, for each candidate neural architecture, the weight estimation is repeated over several \textit{learning cycles}. The average classification performance over such learning cycles is used to determine the \textit{efficacy} of a particular candidate solution as follows:
\begin{small}
\begin{align}
    \label{eq:NetEff}
    \mathcal{E}(\mathcal{X}_i,\mathcal{D}_{test}) & = \displaystyle\frac{1}{cycles} \sum \limits_{k=1}^{cycles} \mathcal{E}_k(\mathcal{X}_i,\mathcal{W}_{i,k}^{\ast},\mathcal{D}_{test})\\
    \label{eq:NetW}
   \mathcal{W}_{i,k}^{\ast} & = \argmin \limits_{\mathcal{W}} \ {\rm L}(\mathcal{X}_i,\mathcal{W},\mathcal{D}_{train})
 \end{align}
\end{small} 
where, $\mathcal{D}_{train}$ and $\mathcal{D}_{test}$ respectively denote the training and testing data; $\rm L(\cdot)$ denotes the \textit{loss} function of the gradient-descent algorithm; and $\mathcal{W}_{i,k}^{\ast}$ gives the \textit{estimated} weight matrix corresponding to $\mathcal{X}_i$ at the $k^{th}$ learning cycle.

\section{Capturing behavior of NASDAQ Index and implications of the COVID pandemic}
\label{sec:Nasdaq_COVID}

Our classification study of the market direction of movement focuses on the NASDAQ Composite, which is a stock market capitalization-weighted index covering  most of the  stocks and specific securities uniquely listed on the NASDAQ stock exchange. The index mainly contains technology companies from the ICT sector and hence is referred to as an indicator for the status quo of the  technology sector.
Its value is calculated by  taking for each share the product of its price (at the previous instance) and a weight according the number of outstanding shares. The composite index is then obtained by summing all products and normalizing according a fixed standard. 
Under accordance with an \emph{adaptive} - vs \emph{efficient} - market `anomalous' changes are identified over its time course; essentially two major anomalies occurred in the recent decennium of the NASDAQ time series which it shares with other stock market indices. The first anomaly in this window is the `December 2018 plunge'  of the stock market; 
for which financial analysts \cite{Isbitts2019} interpret the stock market went from `bull' to `bear' performance because;\\
\indent \emph{``Companies are cutting profit forecasts and trying to temper expectations for earnings growth this year (i.e. 2019) after a big 2018''}, \citep{Moyer2019}.\\
The second anomaly is the `2020 stock market crash' (February  -  April 2020) which resulted from a looming recession but was predicted by inverted yield curves \citep{Li2019}, and followed a synchronized slowdown of the world economy which was identified by the IMF \citep{Gopinath2019}.  Due to the outbreak of the COVID-19 pandemic, with economic shutdowns occurring globally and causing supply chain disruptions and depletions due to panic-buying, subsequently the financial uncertainty impacted the stock markets globally \citep{Karabell2020}. Such anomalous periods in the time series data render the weight learning process of a classifier vulnerable to erroneous signal capturing by the incompatibility of the learning, testing and deployment windows, and degrades the classification accuracy. This incompatibility of time window data can be compared to an inadvertent case of `data poisoning' (see \cite{SteinhardtEtAl_2017} and references herein). Besides these marked trading anomalies, \citet{BuszkoEtAl2021} and \citet{ChandraEtAl2021} report on a long-term anomalous impact of the  COVID-19 pandemic on market trading. This potentially differentiated signal could result in distinct pre- and within-COVID trading tendencies and requires a multi-dataset learning approach. 

\begin{figure}[!t]
    \centering
    \includegraphics[width=0.5\textwidth]{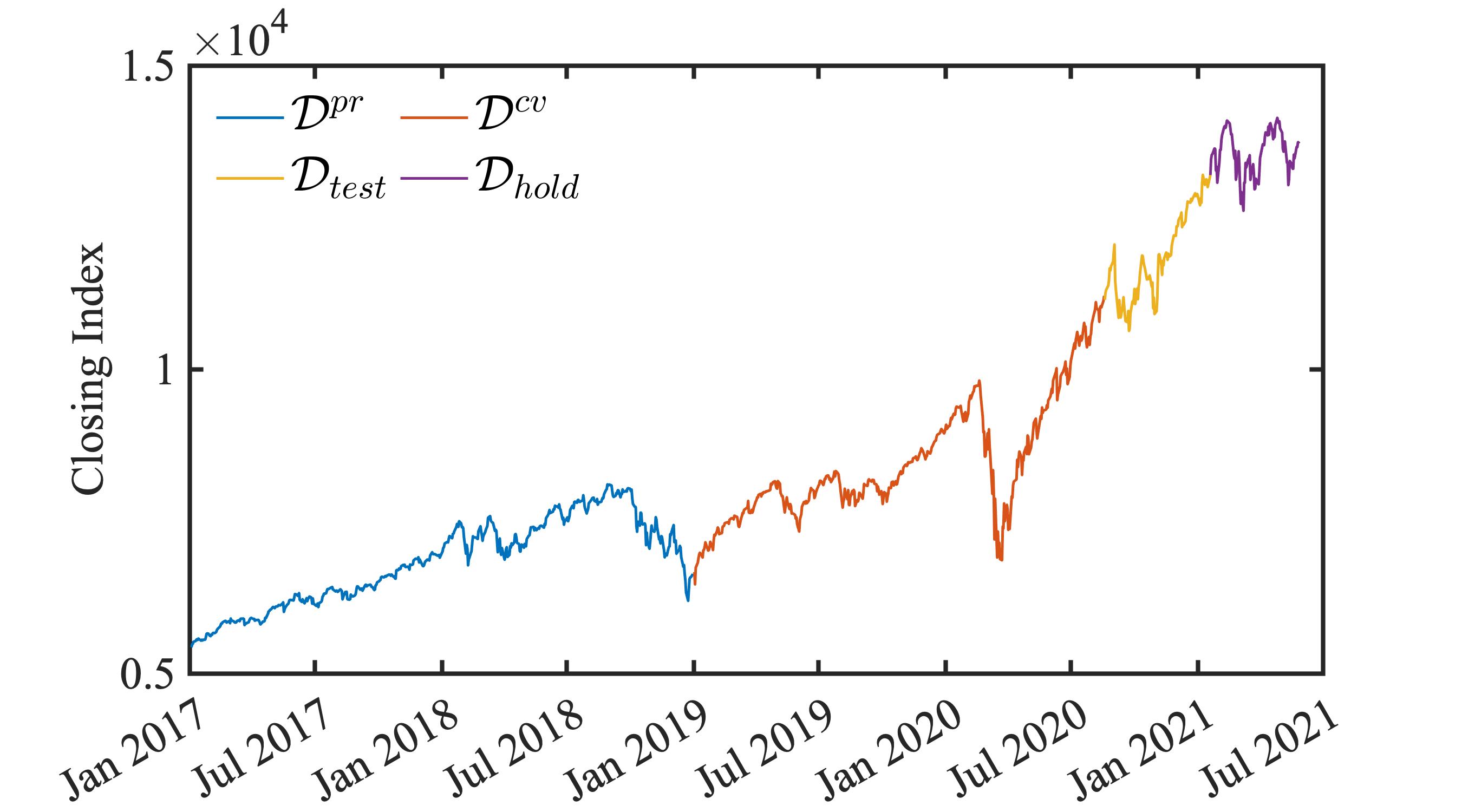}
    \caption{NASDAQ composite closing values segmented into different datasets. $\mathcal{D}^{pr}$:  pre-COVID dataset; $\mathcal{D}^{cv}$: within-COVID \textit{training} dataset; $\mathcal{D}_{test}$ \textit{testing} dataset; and $\mathcal{D}_{hold}$: \textit{hold-out} dataset. The role of these datasets is discussed in detail in Section~\ref{subsec:search}.}
    \label{f:nasdaq}
\end{figure}

In particular, the historical data of NASDAQ index over the approximately past four years is being considered as follows: (1) From January, 2017 to December, 2018: pre-COVID data (denoted as $\mathcal{D}^{pr}$) (2) From January, 2019 to August, 2020: within-COVID \textit{training} data ($\mathcal{D}^{cv}$) (3) From August, 2020 to January, 2021: \textit{testing} data ($\mathcal{D}_{test}$) and (4) From January, 2021 to May, 2021: \textit{hold-out} data ($\mathcal{D}_{hold}$). Fig.~\ref{f:nasdaq} shows the segmentation of the NASDAQ historic data following this scheme. The use of these datasets to train, evolve and test neural architectures will be discussed at length in Section~\ref{subsec:search}.

\subsection{Multi-Dataset Learning: Reconciling Pre- and within-COVID NASDAQ behavior}
\label{subsec:multidataset_learning}

It is expected that distinct trading tendencies may prevail prior to and within the COVID pandemic~\citep{BuszkoEtAl2021,ChandraEtAl2021}, which may hinder the proper tuning of network weights ($\mathcal{W}$). A straight-forward approach to deal with this issue is to discard the Pre-COVID data ($\mathcal{D}^{pr}$) and train neural networks using only using the COVID data ($\mathcal{D}^{cv}$). However, it would be inopportune to forfeit any concordant information due to complete removal of $\mathcal{D}^{pr}$ form the neural design process. Given that it is not trivial to estimate such information, two learning scenario are designed, as discussed in the following:

To test the hypothesis of a distinct NASDAQ trading tendencies and consequent adverse effect on the weight estimation, a \textit{`Full Dataset'} learning scenario (denoted by $\mathcal{LS_F}$) is devised. The efficacy of candidate network architectures under this learning scenario is determined following the steps outlined in Algorithm~\ref{algo:eff_LS_F}. In particular, here, the datasets from both the pre- and within-COVID windows are pooled as the training dataset for weight estimation, see Line~\ref{l:lsf_l1}, Algorithm~\ref{algo:eff_LS_F}. Further, a Penalty function (denoted by $\mathcal{P}$) is introduced to discourage possible convergence to a brute classifier. To this end, a high-penalty is assigned to candidate neural architectures if the corresponding Matthews Correlation Co-efficient (MCC) falls below a pre-fixed threshold, $\epsilon_1$, as follows: 
\begin{small}
\begin{align}
  \label{eq:penalty_full}
  \mathcal{P}(\mathcal{X}_i) & = 5 \times \max \big\{ 0, \ \epsilon_1 - \Phi(\mathcal{X}_i,\mathcal{D}_{test}) \big\}. 
\end{align}
\end{small}
where, $\Phi(\mathcal{X}_i,\mathcal{D}_{test})$ gives the MCC for the $i^{th}$ candidate architecture, $\mathcal{X}_i$ over the test dataset, $\mathcal{D}_{test}$.
\begin{algorithm}[!t]
    \footnotesize
    \SetKwInOut{Input}{Input}
    \SetKwInOut{Output}{Output}
    \SetKwComment{Comment}{*/ \ \ \ }{}
    \Input{Candidate Network, $\mathcal{X}_i \leftarrow \begin{Bmatrix} \mathcal{A}_i, \rm{X}_i \end{Bmatrix}$}
    \Output{Network Efficacy, $\mathcal{E}(\mathcal{X}_i)$, and Penalty, $\mathcal{P}(\mathcal{X}_i)$}
    \BlankLine
    Full training dataset, $\mathcal{D}_{train} \leftarrow \left\{ \mathcal{D}^{pr} \cup \mathcal{D}^{cv} \right\}$\nllabel{l:lsf_l1}
    \BlankLine
    \Comment*[h] {Estimate Network Weights, $\mathcal{W}$}\\
    \BlankLine
    \For{k = 1 to $cycles$}
    {
        \BlankLine
        \Comment*[h] {Scaled conjugate gradient backpropagation~\citep{Moller:1993}}\\
        \BlankLine
        $\mathcal{W}_{i,k}^{\ast} \leftarrow \argmin \limits_{\mathcal{W}} \ {\rm L}(\mathcal{X}_i,\mathcal{W},\mathcal{D}_{train})$\\
        \BlankLine
        \Comment*[h] {Determine Performance over COVID-test data}\\
        $\mathcal{E}_k(\mathcal{X}_i,\mathcal{W}_{i,k}^{\ast},\mathcal{D}_{test}) \leftarrow $ classification error, $\Phi_k(\mathcal{X}_i,\mathcal{W}_{i,k}^{\ast},\mathcal{D}_{test}) \leftarrow $ MCC\\
        \BlankLine
    }
    \BlankLine
    $\mathcal{E}(\mathcal{X}_i,\mathcal{D}_{test}) \leftarrow \displaystyle\frac{1}{cycles} \sum \limits_{k=1}^{cycles} \mathcal{E}_k(\mathcal{X}_i,\mathcal{W}_{i,k}^{\ast},\mathcal{D}_{test})$, \ $\Phi(\mathcal{X}_i,\mathcal{D}_{test}) \leftarrow \displaystyle \frac{1}{cycles} \sum \limits_{k=1}^{cycles} \Phi_k(\mathcal{X}_i,\mathcal{W}_{i,k}^{\ast},\mathcal{D}_{test})$
    \BlankLine
    Network efficacy, $\mathcal{E}(\mathcal{X}_i) \leftarrow \mathcal{E}(\mathcal{X}_i,\mathcal{D}_{test})$\\
    
    Network penalty, $\mathcal{P}(\mathcal{X}_i) \leftarrow 5 \times \Bigg[ \max \big\{ 0, \ \epsilon_1 - \Phi(\mathcal{X}_i,\mathcal{D}_{test}) \big\} \Bigg]$
    
\caption{Estimation of network efficacy under full dataset learning scenario, $\mathcal{LS}_F$}
\label{algo:eff_LS_F}
\end{algorithm}
\begin{algorithm}[!t]
    \footnotesize
    \SetKwInOut{Input}{Input}
    \SetKwInOut{Output}{Output}
    \SetKwComment{Comment}{*/ \ \ \ }{}
    \Input{Candidate Network, $\mathcal{X}_i \leftarrow \begin{Bmatrix} \mathcal{A}_i, \rm{X}_i \end{Bmatrix}$}
    \Output{Network Efficacy, $\mathcal{E}(\mathcal{X}_i)$, and Penalty, $\mathcal{P}(\mathcal{X}_i)$}
    \BlankLine
    Split training dataset, $\mathcal{D}_{train} \leftarrow \mathcal{D}^{cv}$\nllabel{l:lse_l1}
    \BlankLine
    \Comment*[h] {Estimate Network Weights, $\mathcal{W}$}\\
    \BlankLine
    \For{k = 1 to $cycles$}
    {
        \BlankLine
        \Comment*[h] {Scaled conjugate gradient backpropagation~\citep{Moller:1993}}\\
        \BlankLine
        $\mathcal{W}_{i,k}^{\ast} \leftarrow \argmin \limits_{\mathcal{W}} \ {\rm L}(\mathcal{X}_i,\mathcal{W},\mathcal{D}_{train})$\\
        \BlankLine
        \Comment*[h] {Determine Performance over COVID-test data}\\
        $\mathcal{E}_k(\mathcal{X}_i,\mathcal{W}_{i,k}^{\ast},\mathcal{D}_{test}) \leftarrow $ classification error, $\Phi_k(\mathcal{X}_i,\mathcal{W}_{i,k}^{\ast},\mathcal{D}_{test}) \leftarrow $ MCC\\
        \BlankLine
        \Comment*[h] {Determine Performance over Pre-COVID data}\\
         $\mathcal{E}_k(\mathcal{X}_i,\mathcal{W}_{i,k}^{\ast},\mathcal{D}^{pr})\leftarrow $ classification error, $\Phi_k(\mathcal{X}_i,\mathcal{W}_{i,k}^{\ast},\mathcal{D}^{pr})\leftarrow $ MCC\\
        \BlankLine
    }
    \BlankLine
    \Comment*[h] {Network efficacy over COVID-test data}\\
    \BlankLine
    $\mathcal{E}(\mathcal{X}_i,\mathcal{D}_{test}) \leftarrow \displaystyle\frac{1}{cycles} \sum \limits_{k=1}^{cycles} \mathcal{E}_k(\mathcal{X}_i,\mathcal{W}_{i,k}^{\ast},\mathcal{D}_{test})$, \ $\Phi(\mathcal{X}_i,\mathcal{D}_{test}) \leftarrow \displaystyle \frac{1}{cycles} \sum \limits_{k=1}^{cycles} \Phi_k(\mathcal{X}_i,\mathcal{W}_{i,k}^{\ast},\mathcal{D}_{test})$
    \BlankLine
    \Comment*[h] {Network efficacy over Pre-COVID data}\\
    \BlankLine
    $\mathcal{E}(\mathcal{X}_i,\mathcal{D}^{pr}) \leftarrow \displaystyle\frac{1}{cycles} \sum \limits_{k=1}^{cycles} \mathcal{E}_k(\mathcal{X}_i,\mathcal{W}_{i,k}^{\ast},\mathcal{D}^{pr})$, \ $\Phi(\mathcal{X}_i,\mathcal{D}^{pr}) \leftarrow \displaystyle \frac{1}{cycles} \sum \limits_{k=1}^{cycles} \Phi_k(\mathcal{X}_i,\mathcal{W}_{i,k}^{\ast},\mathcal{D}^{pr})$\\
    
    \BlankLine
    Network efficacy, $\mathcal{E}(\mathcal{X}_i) \leftarrow \mathcal{E}(\mathcal{X}_i,\mathcal{D}_{test})$\\
    
    Network penalty, $\mathcal{P}(\mathcal{X}_i) \leftarrow 5 \times \Bigg[ \max \big\{ 0, \ \epsilon_1 - \Phi(\mathcal{X}_i,\mathcal{D}_{test}) \big\} + \max \big\{ 0, \ \epsilon_2 - \Phi(\mathcal{X}_i,\mathcal{D}^{pr}) \big\} + 
    \max \big\{ 0, \ \mathcal{E}(\mathcal{X}_i,\mathcal{D}^{pr}) - \epsilon_3) \big\} \Bigg]$

\caption{Estimation of network efficacy under split dataset learning scenario, $\mathcal{LS}_\varepsilon$}
\label{algo:eff_LS_eps}
\end{algorithm}

In contrast, the second learning scenario indirectly uses the Pre-COVID dataset ($\mathcal{D}^{pr}$), to limit any possible adverse effects on the weight estimation. This scenario is referred to as the \textit{`Split Dataset'} learning scenario and denoted by $\mathcal{LS}_\varepsilon$. The steps involved in this scenario are outlined in Algorithm~\ref{algo:eff_LS_eps}. In particular, the estimation of weight corresponding to candidate architectures is being carried out using only the within-COVID data, $\mathcal{D}^{cv}$, as shown in Line~\ref{l:lse_l1}, Algorithm~\ref{algo:eff_LS_eps}. Further, the Penalty function in this scenario is expanded to accommodate Pre-COVID data ($\mathcal{D}^{pr}$) in the neural architecture search as follows:
\begin{small}
\begin{align}
    \label{eq:penalty_epsilon}
    \mathcal{P}(\mathcal{X}_i) &=  5 \times \Bigg[ \max \Big\{ 0, \ \epsilon_1 - \Phi(\mathcal{X}_i,\mathcal{D}_{test}) \Big\} + \max \Big\{ 0, \ \epsilon_2 - \Phi(\mathcal{X}_i,\mathcal{D}^{pr}) \Big\} + 
    \max \Big\{ 0, \ \mathcal{E}(\mathcal{X}_i,\mathcal{D}^{pr}) - \epsilon_3) \Big\} \Bigg]
\end{align}
\end{small}
where, $\Phi(\mathcal{X}_i,\mathcal{D}_{test})$ and $\Phi(\mathcal{X}_i,\mathcal{D}^{pr})$ respectively give the MCC of $\mathcal{X}_i$ over the test dataset ($\mathcal{D}_{test}$) and the pre-COVID dataset ($\mathcal{D}^{pr}$); and $\mathcal{E}(\mathcal{X}_i,\mathcal{D}^{pr})$ denotes classification error over $\mathcal{D}^{pr}$. It is clear that this formulation of $\mathcal{P}$ encourages the exploration of candidate neural architectures which can maintain classification performance above certain pre-fixed threshold (\textit{controlled by} $\epsilon_1$, $\epsilon_2$ and $\epsilon_3$) in both pre- and within-COVID time periods. In other words, while $\mathcal{D}^{pr}$ is not being used for the weight estimation, it is indirectly used in the search for feature subsets and neural architecture. 

It is worth noting that the goal of the search algorithm (to be discussed in Section~\ref{subsec:2DSpos}) is to minimize the following criterion function $J(\cdot)$, which essentially is a weighted sum of efficacy $(\mathcal{E})$, complexity ($\mathcal{C}$) and Penalty ($\mathcal{P}$):
\begin{small}
\begin{align}
    \label{eq:Jcwa2}
    J(\mathcal{X}_i) & = \theta_\mathcal{E} \ \mathcal{E}(\mathcal{X}_i) \ + \theta_\mathcal{C} \ \mathcal{C}(\mathcal{X}_i) \ + \ \mathcal{P}(\mathcal{X}_i)
\end{align}
\end{small}
It is easy to follow that $\mathcal{P}$ serves as the $\epsilon-$\textit{constraint} to guide the neural architecture search in the desired direction. $\mathcal{P}$ attains the minimum value of $0$ when all the constraints are satisfied, \textit{i.e.}, $\Phi(\mathcal{X}_i,\mathcal{D}_{test}) \geq \epsilon_1$, $\Phi(\mathcal{X}_i,\mathcal{D}^{pr}) \geq \epsilon_2$ and $\mathcal{E}(\mathcal{X}_i,\mathcal{D}^{pr}) \leq \epsilon_3$. Further, since the \textit{worst} value of MCC ($\Phi$) and classification error ($\mathcal{E}$) is respectively $-1$ and $1$, the maximum value of $\mathcal{P}$ is determined to be $(\epsilon_1+1)$ for $\mathcal{LS_F}$ and $(\epsilon_1+\epsilon_2-\epsilon_3+3)$ for $\mathcal{LS}_\varepsilon$.

To ensure that a search does not converge to a \textit{brute} classifier, the threshold values for MCC are set $0.2$, \textit{i.e.}, $\epsilon_1 = \epsilon_2 = 0.2$. Given that the lowest acceptable accuracy threshold for stock market prediction is around $56\%$, the classification error threshold is set to $\epsilon_3 = 0.44$.

To summarize, the full learning scenario $\mathcal{LS_F}$ has the advantage of more data samples for weight training with the potential risk of inadvertent \textit{data-poising} (subset incompatibility) due to possibly distinct market behavior prior to COVID. The second scenario $\mathcal{LS}_\varepsilon$, reduces the risks stemming from  of data incompatibility in the weight estimation while integrating $\mathcal{D}^{pr}$ through the $\epsilon-$constraint framework. The performance of neural architectures evolved under these two learning scenarios will be discussed in Section~\ref{subsec:comparison_of_learning_scenarios}.

\section{Search for the Optimal Neural Architecture: Two-Dimensional Swarms (2DS)}
\label{sec:search_opt_NN_top}

The exhaustive search as well as most exact methods are often not suitable for the Extended Neural Architecture Search (ENAS) due to the associated exponential and intractable search space (see Section~\ref{sec:NAS}). The population based search heuristics such as Genetic Algorithm (GA) and Binary Particle Swarms (BPSO) are often found to be effective on such problems~\citep{Xue:Zhang:2016}. Further, one of the key aspects of ENAS, which is driven by the \textit{principle-of-parsimony} or \textit{Occam's Razor}, is to judiciously estimate \textit{complexity/sparsity} of candidate solutions for the dataset under consideration. This study, therefore, proposes the extension of population based search heuristic, Two-Dimensional Swarms (2DS), which explicitly focuses on the \textit{sparsity} of the candidate solutions~\citep{Hafiz:Swain:2018,Hafiz:Swain:SI2DUPSO:2019,Hafiz:Swain:ASOC:2019,Hafiz:Swain:CEC:2018}.

2DS was originally developed by the authors for the feature selection problem in~\citep{Hafiz:Swain:2018,Hafiz:Swain:ASOC:2019} and subsequently extended for regressor/term selection in nonlinear system identification in~\citep{Hafiz:Swain:CEC:2018,Hafiz:Swain:SI2DUPSO:2019}. In this study, we build upon 2DS framework to identify \textit{parsimonious} and \textit{efficacious} combination of neural architecture and feature subset, as will be discussed in the following subsections.

\begin{figure}[!t]
\centering
\begin{adjustbox}{width=0.55\textwidth}
\begin{tikzpicture}[
node distance=0pt,
 start chain = A going right,
    X/.style = {rectangle, draw,
                minimum width=2ex, minimum height=3ex,
                outer sep=0pt, on chain},
    X1/.style = {
                minimum width=2ex, minimum height=3ex,
                outer sep=1.5pt, on chain},
    B/.style = {decorate,
                decoration={brace, amplitude=5pt,
                pre=moveto,pre length=3pt,post=moveto,post length=3pt,
                raise=2mm,
                            #1}, 
                thick},
    B/.default=mirror, 
                        ]
    \foreach \i in {0,1,1,$\dots$,0,0,1}
    \node[X] {\i};
    
    \foreach \i in {$\dots$}
    \node[X1] {\i};
    
    \foreach \i in {1,1,0,$\dots$,1,0,0}
    \node[X] {\i};
    
    \foreach \i in {1,0,1,$\dots$,0,0,1}
    \node[X] {\i};

\draw[densely dotted] 
([yshift=4em] A-16.north west) -- ( [yshift=-4em] A-16.south west);

\draw[B] ( A-1.south west) -- node[below=4mm,text width=3cm,align=center] {hidden layer-1\\ ($n_{bits}$)} ( A-7.south east);

\draw[B] (A-9.south west) -- node[below=4mm,text width=5cm,align=center] {hidden layer-$n_\ell$\\ ($n_{bits}$)} (A-15.south east);

\draw[B=] (A-16.north west) -- node[above=4mm,text width=3cm,align=center] {feature encoding \\ ($n_f-$bits)} (A-22.north east);

\draw[B=] (A-1.north west) -- node[above=4mm,text width=5cm,align=center] {architecture encoding\\ $(n_\ell \times n_{bits})$} (A-15.north east);

\end{tikzpicture}
\end{adjustbox}

\caption{Encoding of a candidate solution into an `$n$'-dimensional binary string, where $n=n_f + (n_\ell \times n_{bits})$.Each hidden layer is encoded by `$n_{bits}$' binary bits. Accordingly entire architecture is encoded by $(n_\ell \times n_{bits})$ bits. The feature subset is encoded by `$n_f$' bits.}
\label{f:binary}
\end{figure}



\subsection{Solution Encoding}
\label{subsec:encoding}

In this study, a candidate solution is encoded into a two part binary string, $\mathcal{B} = \begin{Bmatrix} B_\mathcal{A}, & B_{\rm X} \end{Bmatrix}$. The first part of such string ($B_\mathcal{A}$) encodes the neural architecture and the second part ($B_{\rm X}$) stores the feature subset, as discussed in the following:

The information about each hidden layer is encoded in an $n_{bits}-$ binary tuple. All but the last bit of this tuple encodes the \textit{size} ($s$) or number hidden neurons by \textit{decimal-to-binary} conversion. The final bit of such tuple is used to encode selection of a particular activation function, $f \in \mathcal{F}$. In this study, we limit the choice of activation functions to two as follows: $\mathcal{F} = \left\{sigmoid, \ tanh \right\}$. Consequently, a single bit binary encoding is sufficient to represent the selection of activation for a particular hidden layer. 

Given that the neural architecture can have maximum $n_\ell$ hidden layers, the entire architecture can be encoded by a total of $(n_{bits} \times n_\ell)$ binary bits, as follows:
\begin{small}
\begin{align*}
    {B}_\mathcal{A} = \begin{Bmatrix} \beta_{1}, & \beta_{2}, & \dots & \beta_{(n_{bits} \times n_\ell)} \end{Bmatrix}, \quad \beta_{j} \in \{0,1\}, \ \ j=1,2,\dots (n_{bits} \times n_\ell)
\end{align*}
\end{small}
This is further explained by the encoding scheme depiction in Fig.~\ref{f:binary}. 

Next, the encoding of the feature subset is relatively simple: feature subset is encoded by an $n_f-bits$ binary string, wherein each bit represents that whether the corresponding feature is to be included, as follows:
\begin{small}
\begin{align*}
    {B}_{\rm X} = \begin{Bmatrix} \beta_{1} & \beta_{2} & \dots & \beta_{n_f} \end{Bmatrix} \quad \beta_{j} \in \{0,1\}, \ \ j=1,2,\dots n_f
\end{align*}
\end{small}
The $j^{th}$ bit $\beta_{j} \in B_{\rm X}$ is set to `$1$' provided that the $j^{th}$ feature ($x_j \in {\rm {X}_{full}}$) is included in into the feature subset ${\rm X}$.

In the final step, the neural architecture encoding ($B_\mathcal{A}$) and feature subset encoding ($B_{\rm X}$) are combined to derive the final $n-$dimensional bit string `$\mathcal{B}$' corresponding to the candidate solution `$\mathcal{X}$', as follows:
\begin{small}
\begin{align} 
\label{eq:beta}
    \mathcal{B} & = \begin{Bmatrix} B_\mathcal{A}, & B_{\rm X} \end{Bmatrix} = \begin{Bmatrix} \beta_{1} & \beta_{2} & \dots & \beta_{(n_{bits} \times n_\ell)}, & \dots & \beta_{n} \end{Bmatrix}
\end{align}
\end{small}
where, $n = (n_{bits} \times n_\ell) \ + \ n_f$. The first $(n_{bits} \times n_\ell)$ bits of $\mathcal{B}$ encodes the neural architecture $\mathcal{A}$ and the remaining bits encodes the feature subset $\rm X$. The entire encoding scheme is depicted in Fig.~\ref{f:binary} for ease of interpretation. 

\begin{algorithm}[!t]
    \footnotesize
    \SetKwInOut{Input}{Input}
    \SetKwInOut{Output}{Output}
    \SetKwComment{Comment}{*/ \ \ \ }{}
    \Input{Search Agent, $\mathcal{B}_i = \begin{Bmatrix} \beta_{i,1}, & \beta_{i,2}, & \dots, & \beta_{i,n} \end{Bmatrix}$}
    \Output{Criterion Function, $J(\mathcal{X}_i)$}
    \BlankLine
    \Comment*[h] {Decode the neural architecture, $\mathcal{A}_i$}\\
    \BlankLine
    $\mathcal{A}_i \leftarrow \emptyset$\nllabel{l:crf1}\\
    \For{j = 1 to $n_\ell$} 
        { \BlankLine
           \Comment*[h] {Layer Size}\\
           $\displaystyle s_j \leftarrow \sum \limits_{p=1}^{(n_{bits}-1)} \beta_{i,p} \times 2^{(n_{bits} -p -1)}$ \Comment*[h] {\textit{binary-to-decimal}}\\
           \BlankLine
           \Comment*[h] {Activation Function}\\
           \uIf{$\beta_{i,n_{bits}}$ = 1}
           { $f^j \leftarrow \textit{sigmoid} $}
           \Else{$f^j \leftarrow \tanh $}
           \BlankLine
           $\mathcal{A}_i \leftarrow \left\{\mathcal{A}_i \cup (s^j,f^j) \right\}, \qquad \qquad$  $\mathcal{B}_i \leftarrow \mathcal{B}_i \setminus \begin{Bmatrix} \beta_{i,1}, & \beta_{i,2}, & \dots, & \beta_{i,n_{bits}} \end{Bmatrix}$
        } 
    \BlankLine \nllabel{l:crf2}
    \BlankLine
    \Comment*[h] {Decode the feature subset, ${\rm X}_i$}\\
    \BlankLine
    ${\rm X}_i \leftarrow \emptyset$\\
    \For{j = 1 to $n_f$} 
        { \BlankLine
           \If{$\beta_{i,j}=1$}
           {\BlankLine
             ${\rm X}_i \leftarrow \{ {\rm X}_i \cup x_j \}$  \Comment*[h] {add the $j^{th}$ feature}\\
            \BlankLine
           }
          \BlankLine
        }
    \BlankLine
    \Comment*[h] {Candidate Network}\\
    \BlankLine
    $\mathcal{X}_i \leftarrow \begin{Bmatrix} \mathcal{A}_i, \rm{X}_i \end{Bmatrix}$ \nllabel{l:crf3}\\
    \BlankLine
    \Comment*[h] {Estimation of Network Performance}\\
    \BlankLine
    Estimate network efficacy, $\mathcal{E}(\mathcal{X}_i)$, and Penalty, $\mathcal{P}(\mathcal{X}_i)$ following the step outlined in Algorithm~\ref{algo:eff_LS_F} or~\ref{algo:eff_LS_eps} \\
    \BlankLine
    \Comment*[h] {Determine Network Sparsity}\\
    \BlankLine
    $\mathcal{C}(\mathcal{X}_i) \leftarrow \displaystyle \frac{1}{3} \Big\{ \frac{\#(X_i)}{n_f} + \frac{\# \Big( \begin{Bmatrix} (s_k,f_k) \ | \ s_k \neq 0 \end{Bmatrix} \Big)}{n_\ell} + \sum \limits_{k=1}^{n_\ell} \frac{s^k}{s^{max}} \Big\}$\\
    \BlankLine
    $J(\mathcal{X}_i) \leftarrow  \theta_\mathcal{E} \ \mathcal{E}(\mathcal{X}_i) \ + \theta_\mathcal{C} \ \mathcal{C}(\mathcal{X}_i) \ + \ \mathcal{P}$
    
\caption{Evaluation of the Multi-Objective Criterion Function, $J(\cdotp)$}
\label{al:crf}
\end{algorithm}

It is worth noting that during the search process each candidate solution is encoded in the binary string representation given by~(\ref{eq:beta}). Further, \textit{decoding} steps required to extract the neural architecture ($\mathcal{A}$) and the feature subset ($\rm X$) corresponding to the binary string ($\mathcal{B}$) are outlined in Lines~\ref{l:crf1}-\ref{l:crf3}, Algorithm~\ref{al:crf}.

\subsection{Two-Dimensional Swarms: Overall Framework}
\label{subsec:2DSOverall}

Before we discuss the methodology of the Two-Dimensional (2D) learning, it is pertinent to briefly discuss the overall steps involved in 2DS. 

2DS is developed under the framework of \textit{particle swarm theory}~\citep{Kennedy:Eberhart:1995,Kennedy:Mendes:2002}, and it essentially imitates the \textit{social co-operation} for foraging. The central idea to such imitation is to \textit{`learn'} from peers, \textit{i.e.}, each search agent uses the information extracted from its peers to adjust its trajectory on the solution landscape. 2DS retains the mechanisms of conventional particle swarms to imitate social co-operation, while proposing completely new \textit{learning framework} as will be discussed in Section~\ref{subsec:2DSLearn}-\ref{subsec:2DSpos}.

2DS begins with a \textit{swarm} of pre-fixed number of search agents or \textit{particles}. Each particle has the following three attributes which are updated iteratively throughout the search process: \textit{position} (denoted by $\mathcal{B}$), \textit{velocity} ($\mathcal{V}$) and \textit{memory}. The position, $\mathcal{B}$, encodes a candidate solution, as discussed in Section~\ref{subsec:encoding}. The velocity of the particle stores the \textit{selection likelihoods} based on which $\mathcal{B}$ is determined in each iteration (to be discussed in Section~\ref{subsec:2DSpos}).

In addition, as the \textit{memory} attribute, each particle stores the best solution hitherto discovered by itself as the \textit{personal best position}, $\mathcal{B}_P$. To mimic the social co-operation, each particle shares the personal best position with its peers. This sharing mechanisms allows the discovery of best position (candidate solution) in a \textit{neighborhood} of the particle (\textit{neighborhood best}, $\mathcal{B}_N$) as well in the entire swarm (\textit{global best}, $\mathcal{B}_G$).
Each particle iteratively updates and stores the \textit{best} positions ($\mathcal{B}_G$, $\mathcal{B}_N$ and $\mathcal{B}_P$). A detailed treatment on information sharing with peers, different neighborhood topologies and their implications on the search performance can be found in~\cite{Kennedy:Mendes:2002}.

\begin{algorithm}[!t]
    \footnotesize
    \SetKwInOut{Input}{Input}
    \SetKwInOut{Output}{Output}
    \SetKwComment{Comment}{*/ \ \ \ }{}
    \Input{Supervised Learning data; $\mathcal{D}^{cv}_{train}$, $\mathcal{D}^{cv}_{test}$ and $\mathcal{D}^{pr}$; Search parameters: $u_f$ and $RG$}
    \Output{Optimal Solution, $\mathcal{X}^\ast \in \Omega$}
    \algorithmfootnote{$^\dagger$The \textit{ring-topology} (see~\citet{Kennedy:Mendes:2002}) is used to find neighborhood best, $\mathcal{B}_N$; `$pbestval_i^t$' denotes the criterion function, $J(\cdot)$, corresponding to the personal best of the $i^{th}$ particle, $\mathcal{B}_P^i$, at iteration $t$.}
    \BlankLine
    Randomly initialize the swarm of `$ps$' number of particles, $\{ \mathcal{B}_1 \dots \mathcal{B}_{ps} \}$ \\
    Initialize the velocity of each particle, $\mathcal{V} \in \mathbb{R}^{2\times n}$, by uniformly distributed random numbers in [0,1] \\
    \BlankLine
    Evaluate criterion function ($J$) of each particle (see Algorithm~\ref{al:crf})\\
    Determine the learning exemplars ($\mathcal{B}_P$, $\mathcal{B}_G$ and $\mathcal{B}_N$), see~\citep{Kennedy:Mendes:2002} \\
    \BlankLine
    \For{t = 1 to iterations}
        { 
            \BlankLine
            \Comment*[h]{Swarm Update}\\
            \BlankLine
            \For{i = 1 to ps} 
            {
                \BlankLine
                \Comment*[h]{Re-initialize Particles beyond Refresh Gap}\\
                \BlankLine
                \If{$ count_i \geq RG$\nllabel{line:RG1}}
                    {Re-initialize the velocity by uniformly distributed random numbers\\
                     $V_i \leftarrow \mathbb{R}^{2\times n}$, $V_i\in [0,1]$\\
                     $count_i \leftarrow 0$}
                \BlankLine \nllabel{line:RG2}
                Evaluate the learning sets, $\mathcal{L}_{P}$, $\mathcal{L}_{G}$, $\mathcal{L}_{N}$ and $\mathcal{L}_{i}$, as per Algorithm~\ref{al:learn}\\
                Update the velocity of the $i^{th}$ particle as per~(\ref{eq:velupdate})\\
                Update the position of the $i^{th}$ particle following Algorithm \ref{al:pos}
            }
            \BlankLine
            Evaluate the swarm fitness, $\overrightarrow{J^t}$, as per Algorithm~\ref{al:crf}\\
            \BlankLine
            Update learning exemplars, ($\mathcal{B}_P$, $\mathcal{B}_G$ and $\mathcal{B}_N$), see~\citep{Kennedy:Mendes:2002}\\             
            \BlankLine
            \Comment*[h]{Stagnation Check for Refresh Gap}\\ 
            \BlankLine
            \For{i = 1 to ps\nllabel{line:RG3}} 
                { \If{$pbestval_i^{ \ t} \geq pbestval_i^{ \ t-1}$}
                        {$count_i \leftarrow count_i+1$}
                } 
            \BlankLine\nllabel{line:RG4}
        }
\caption{Pseudo code of 2DS algorithm$^\dagger$}
\label{al:2DS}
\end{algorithm}

It is worth noting that the key search mechanism here is to extract the \textit{`beneficial'} information from the \textit{best positions}, \textit{e.g.}, \textit{relevant features, activation function and neural topological information}. Such information is subsequently used to update the selection likelihoods which are stored in the particle velocity, $\mathcal{V}$. This information extraction process is referred to as \textit{`learning'} in 2DS, which will be discussed in Section~\ref{subsec:2DSvel}. Further, the best positions are referred to as \textit{learning exemplars} due to their significant influence on the search process.

To summarize, the overall procedures involved in each iteration of 2DS can broadly be categorized in the following steps:
\begin{itemize}
    \item \textit{Learning}: the information from the learning exemplars is extracted and stored in \textit{learning sets}
    \smallskip
    \item \textit{Velocity Update:} the selection likelihoods are updated using the learning sets
    \smallskip
    \item \textit{Position Update:} the positions (candidate solutions) are updated using the selection likelihoods 
    \smallskip
    \item \textit{Exemplar Update:} learning exemplars ($\mathcal{B}_P$, $\mathcal{B}_G$ and $\mathcal{B}_N$) are updated
\end{itemize}

These steps are outlined Algorithm~\ref{al:2DS}. Note that 2DS also includes a \textit{restart mechanism}, to prevent swarm \textit{stagnation} (see~\citep{Hafiz:Swain:2018,Hafiz:Swain:SI2DUPSO:2019}, for details). To detect stagnation, the personal best position ($\mathcal{B}_P$) of each particle is monitored (Lines \ref{line:RG3}-\ref{line:RG4}, Algorithm~\ref{al:2DS}), if it does not improve for several iterations, the particle is deemed to be \textit{stagnated}. Subsequently, the selection likelihoods of stagnated particle are randomly re-initialized to rejuvenate the search process, see Lines \ref{line:RG1}-\ref{line:RG2}, Algorithm~\ref{al:2DS}.

\subsection{Learning Methodology of 2DS}
\label{subsec:2DSLearn}

The Occam's Razor often plays a major role in many topological and combinatorial optimization problems associated with data-driven modeling. For instance, in feature selection, many \textit{relevant/significant} features can be \textit{redundant}. The set of optimal features, thus, can be subset of all relevant features. Focusing part of the search process on \textit{parsimonious} solutions is therefore likely to improve the search performance. This is further corroborated by the results of our earlier investigations in~\citep{Hafiz:Swain:2018,Hafiz:Swain:ASOC:2019,Hafiz:Swain:SI2DUPSO:2019}, in which the search algorithms with \textit{unitary} search focus only on \textit{relevant} variables (\textit{e.g.}, GA, BPSO and ACO) were found to be ineffectual in removing \textit{redundant} variables. By introducing \textit{solution sparsity} as the other search dimension, the search efforts can be directed towards \textit{efficacious} and \textit{parsimonious} solutions, which can effectively remove \textit{redundant} variables. This has been the motivation behind the Two-Dimensional Learning methodology.

In Two-Dimensional Swarms (2DS), each particle focuses on two distinct and independent \textit{learning} dimensions, as the name suggests. The first dimension is dedicated to learn about \textit{solution sparsity}. To this end, the sparsity is estimated via the \textit{cardinality} of non-zero bits in the corresponding binary encoding, $\mathcal{B}$ (see Section~\ref{subsec:encoding}). The goal of this dimension is to evolve selection likelihoods of distinct solution sparsity by controlling the cardinality of the corresponding binary encoding. On the other hand, the second learning dimension focuses on the \textit{solution efficacy}. To this end, the second dimension evolves the selection likelihoods of individual bits of $\mathcal{B}$, which can encode a \textit{feature}, \textit{layer size} or an \textit{activation function}. 

This can further be illustrated by considering the velocity ($\mathcal{V}$) of a particular particle. Since the selection likelihoods are stored in the velocity of each particle, the objectives of each learning dimension can be examined as follows: 
\begin{small}
\begin{align}
\label{eq:vel}
\mathcal{V}_i & =\begin{bmatrix} v_{11}^i & v_{12}^i & \dots & v_{1n}^i \\        
                     v_{21}^i & v_{22}^i & \dots & v_{2n}^i \end{bmatrix}, \ \text{where, \ } v \in \mathbb{R}
\end{align}
\end{small}
where, $\mathcal{V}_i$ denotes the velocity of the $i^{th}$ particle. The first row of $\mathcal{V}_i$ stores the selection likelihoods for \textit{cardinality} of $\mathcal{B}_i$. For instance, `$v_{1,j}^i$' gives the likelihood that cardinality of $\mathcal{B}_i$ will be equal to `$j$', \textit{i.e.}, $p\Big(\#(\mathcal{B}_i)=j \Big) = v_{1,j}^i, \ j\in[1,n]$. In contrast, the second row of $\mathcal{V}_i$ gives the selection likelihoods for individual bits of $\mathcal{B}_i$. For example, `$v_{2,j}^i$' gives the likelihood that the $j^{th}$ bit of $\mathcal{B}_i$ is set to `$1$', \textit{i.e.}, $p\Big( \beta_{i,j} = 1 \Big) = v_{2,j}^i, \ j\in[1,n]$.

The selection likelihoods of each particle are updated in each iteration using the \textit{learning sets} derived from the learning exemplars ($\mathcal{B}_N$, $\mathcal{B}_G$ and $\mathcal{B}_P$) and a simple velocity update rule, which will be discussed in Section~\ref{subsec:2DSvel}. Finally, the new candidate solutions are explored through combined use of \textit{cardinality} and \textit{bit} selection likelihoods. This procedure is discussed in Section~\ref{subsec:2DSpos}.

\subsection{Velocity Update: Evolving Selection Likelihoods}
\label{subsec:2DSvel}

The learning exemplars like \textit{personal} ($\mathcal{B}_P$), \textit{neighborhood} ($\mathcal{B}_N$) and \textit{global} ($\mathcal{B}_G$) best positions represents the best solution found hitherto by a particle, its neighborhood and the swarm, respectively. The first step to update the selection likelihoods is, therefore, to extract the \textit{learning} from such exemplars. The \textit{learning}, in the context of neural architecture search, represents information about `beneficial' \textit{features}, \textit{layer sizes} and \textit{activation function} of the hidden layers as well as \textit{solution complexity}. Accordingly, each exemplar is examined from two distinct perspectives: 1) Cardinality: \textit{how many non-zero bits have been included in the exemplar?} and 2) Significant Bits: \textit{which bits have been selected in the exemplar?} This can be accomplished with few bit-wise logical operations since both the learning exemplar and the particle position are $n-$dimensional binary strings, as follows:

Let the following $n-$dimensional binary string, $\mathcal{B}_\lambda$, represent a particular learning exemplar:
\begin{small}
\begin{align}
    \label{eq:Blambda}
    \mathcal{B}_\lambda = \begin{bmatrix} \beta_{\lambda,1}, & \beta_{\lambda,2}, & \dots, & \beta_{\lambda,n} \end{bmatrix}
\end{align}
\end{small}
The cardinality of $\mathcal{B}_\lambda$ then can easily be determined as follows: $\# (\mathcal{B}_\lambda) = \sum \limits_{k=1}^{n} \beta_{\lambda,k}$. 

Next the non-zero bits of the learning exemplar ($\mathcal{B}_\lambda$) and the particle position ($\mathcal{B}$) are compared by a simple bit-wise logical AND operation as follows: $\mathcal{B}_\lambda \wedge \overline{\mathcal{B}}$, where $\overline{\mathcal{B}}$ denotes logical compliment of $\mathcal{B}$. The objective here is to identify the features, layer sizes and activation functions which have been selected in $\mathcal{B}_\lambda$ but are not included in $\mathcal{B}$. 

The overall steps involved in the learning process are outlined in Algorithm~\ref{al:learn}. As outlined in Line~\ref{line:ls5}, Algorithm~\ref{al:learn}, the learning outcomes are encoded into a binary \textit{learning matrix}, $\mathcal{L}$ of size ($2 \times n$), as follows: 
\begin{small}
\begin{align*}
\mathcal{L} & = \begin{bmatrix} l_{1,1} & l_{1,2} & \dots & l_{1,n} \\                     
                     l_{2,1} & l_{2,2} & \dots & l_{2,n} \end{bmatrix}, \ \text{where, \ } l \in \{0,1\}
\end{align*}
\end{small}
\begin{algorithm}[!t]
    \footnotesize
    \SetKwInOut{Input}{Input}
    \SetKwInOut{Output}{Output}
    \SetKwComment{Comment}{*/ \ \ \ }{}
    \Input{Learning Exemplar, $\mathcal{B}_\lambda = \begin{bmatrix} \beta_{\lambda,1}, & \dots, & \beta_{\lambda,n} \end{bmatrix}$; Particle Position, $\mathcal{B}_i = \begin{bmatrix} \beta_{i,1}, & \dots, & \beta_{i,n} \end{bmatrix}$}
    \Output{Learning Sets: $\mathcal{L}_\lambda$ and $\mathcal{L}_i$}
    \algorithmfootnote{`$\wedge$' denotes bit-wise logical `AND' operation. `$\overline{\mathcal{B}_i}$' denotes logical complement of `$\mathcal{B}_i$'}
    \BlankLine
    \Comment*[h] {Learning for Subset Cardinality}\\
    Initialize cardinality learning vectors, \textit{i.e.}, $\varphi_{\lambda} \leftarrow \{ 0, \ 0 \dots 0 \}, \quad  \varphi_{i} \leftarrow \{ 0, \ 0 \dots 0 \}$ \nllabel{line:ls1}
    \BlankLine
    $ \varphi_{\lambda,j} \leftarrow 1 \ \Big| \ j =\sum \limits_{k=1}^{n} \beta_{\lambda,k} , \ j \in [1,n]$; $\qquad$ $\varphi_{i,j} \leftarrow 1 \ \Big| \ j = \sum \limits_{k=1}^{n} \beta_{i,k}, \ j \in[1,n]$ \\
    \BlankLine
    \BlankLine
    \Comment*[h] {Learning for Bits}\\  
    \BlankLine
    Evaluate Bit Learning Vector: $\psi_{\Lambda} \leftarrow \{ \mathcal{B}_\lambda \wedge \overline{\mathcal{B}_i} \}$ and $\psi_{i} \leftarrow \mathcal{B}_i$ \\ \nllabel{line:ls4}
    $\mathcal{L}_{\lambda} \leftarrow \begin{bmatrix} \varphi_{\lambda} \\ \psi_{\lambda} \end{bmatrix}$ and $\mathcal{L}_{i} \leftarrow \begin{bmatrix} \varphi_{i} \\ \psi_{i} \end{bmatrix}$ \Comment*[h] {Complete Learning Matrices}\nllabel{line:ls5}
    \BlankLine
\caption{Evaluation of the learning sets}
\label{al:learn}
\end{algorithm}

Once the learning matrix corresponding to each exemplar is determined, the selection likelihoods are updated through the following velocity update rule:
\begin{small}
\begin{align}
\label{eq:velupdate} 
\mathcal{V}_{i}^{t+1}  & = \mathcal{V}_{i}^t + (r_1 \times \mathcal{L}_{P}) + (u_f r_2 \times \mathcal{L}_{G}) + (r_3 (1-u_f) \times \mathcal{L}_{N}) + (\Delta_{i} \times \mathcal{L}_{i})\\
\label{eq:fitfeedback}
\text{where,} \    \Delta_i & = \begin{cases}
                \displaystyle\frac{max(\overrightarrow{J^{t}}) - J_{i}^{t}}{max(\overrightarrow{J^{t}}) - min(\overrightarrow{J^{t}})}, &   if \quad J_{i}^{t}<J_{i}^{t-1}\\
                0 , &   otherwise
                 \end{cases} \ \
\end{align}
\end{small}
where, $\mathcal{V}_{i}^t$ denotes the velocity of the $i^{th}$ particle at iteration-$t$; $\mathcal{L}_{P}$, $\mathcal{L}_{G}$, $\mathcal{L}_{N}$ and $\mathcal{L}_{i}$ respectively denote the learning matrix derived from $\mathcal{B}_P$, $\mathcal{B}_G$, $\mathcal{B}_N$ and $\mathcal{B}_i$; $r_1,r_2,r_3 \in \mathbb{R}^{2\times n}$ are uniformly distributed random numbers in $[0,2]$; $u_f \in [0,1]$ is the tuning parameter of the algorithm; `$J_{i}^{t}$' gives the \textit{criterion function} of the $i^{th}$ particle at iteration-$t$; `$\overrightarrow{J^t}$' stores the \textit{criterion function} of the entire swarm at iteration-$t$.

A detailed discussion on the influence of control parameters ( \textit{e.g.}, $u_f$ and $\Delta_i$), their influence on the search process and the rationale behind the selection of the velocity update rule can be found in~\citep{Hafiz:Swain:2018,Hafiz:Swain:ASOC:2019,Hafiz:Swain:SI2DUPSO:2019}. 


\subsection{Position Update: Explorations of New Solutions}
\label{subsec:2DSpos}

The new positions of the particles, and thereby the new candidate solutions, are determined using the combined use of \textit{cardinality} and \textit{bit} selection likelihoods. To this end, the cardinality of the new position, $\#(\mathcal{B}_i)$, is determined first, through the \textit{roulette wheel selection} on the cardinality likelihoods, \textit{i.e.}, $\begin{bmatrix} v_{1,1}^i & v_{1,2}^i & \dots v_{1,n}^i \end{bmatrix}$. This procedure is outlined in Lines~\ref{line:pos1}-\ref{line:pos2}, Algorithm~\ref{al:pos}. 

Next, the \textit{bits} are \textit{ranked} in the descending order of their corresponding selection likelihoods, \textit{i.e.}, $\begin{bmatrix} v_{2,1}^i & v_{2,2}^i & \dots v_{2,n}^i \end{bmatrix}$. Subsequently, the first $\#(\mathcal{B}_i)$ number of bits are selected in the new position. These steps are shown in Lines~\ref{line:pos3}-\ref{line:pos4}, Algorithm~\ref{al:pos}.

The criteria function, $J(\cdot)$, corresponding to each new position is determined following the steps outlined in Algorithm~\ref{al:crf}.


\begin{algorithm}[!t]
    \footnotesize
    \SetKwInOut{Input}{Input}
    \SetKwInOut{Output}{Output}
    \SetKwComment{Comment}{*/ \ \ \ }{}
    \Input{velocity of the $i^{th}$ particle: $\mathcal{V}_i$}
    \Output{position of the $i^{th}$ particle: $\mathcal{B}_i = \begin{bmatrix} \beta_{i,1}, & \beta_{i,2}, & \dots, & \beta_{i,n} \end{bmatrix}$ }
    \BlankLine
    \Comment*[h] {Roulette wheel selection of the cardinality, $\#(\mathcal{B}_i)$}\\
    Determine selection probabilities, $p_k \leftarrow {\Sigma_k}/{\Sigma_n}$, where, $\Sigma_k \leftarrow \sum \limits_{j=1}^{k} v_{1,j}^i, \ \  k = 1\dots n$ \nllabel{line:pos1}\\
    \BlankLine
    $ \#(\mathcal{B}_i) \leftarrow \Big \{ k \ | \ p_{k-1} < r < p_k,  \ k \in [1,n] \Big \}$, where $r \in [0,1]$ is a uniform random number \nllabel{line:pos2}\\
    \BlankLine
    \Comment*[h]{Selection of the bits}\\
    \textit{`rank'} the bits in the descending order of their \textit{likelihoods}, \textit{i.e.}, $v^i_{2,rank(1)} > v^i_{2,rank(2)} > \dots > v^i_{2,rank(n)}$\nllabel{line:pos3}\\
    \BlankLine
    \For{j = 1 to $n$} 
        {  \BlankLine 
           \uIf{$rank(j) \leq \#(\mathcal{B}_i)$}
            {$\beta_{i,j} \leftarrow 1$}
          \Else{$\beta_{i,j} \leftarrow 0$}
        } \nllabel{line:pos4}
\caption{The position update of the $i^{th}$ particle in 2DS}
\label{al:pos}
\end{algorithm}

\section{Results}
\label{sec:results}

Before we discuss the results, it is pertinent to briefly revisit the main investigative focus of this study. In particular, the neural architecture search for the day-ahead prediction of NASDAQ movement is addressed with the following two objectives:
\smallskip
\begin{itemize}
    \item \textbf{Learning Scenario and \textit{Data incompatibility}}: The objective here is to determine the best way to integrate the Pre-COVID data, $\mathcal{D}^{pr}$, in the evolution of efficacious neural architectures. To this end, two learning scenarios are being considered (see Section~\ref{subsec:multidataset_learning}) and their implications on neural architecture search and the potential inadvertent \textit{`data-poisoning'} effects are discussed in Section~\ref{subsec:comparison_of_learning_scenarios}.
    \smallskip
    \item \textbf{Neural Architecture Search}: The other objective is to highlight the benefits of simultaneous evolution of feature subset and neural topology (see Section~\ref{sec:NAS}). To this end, empirical rules of neural architectures selection are considered as the \textit{baseline} approach, and their performance is discussed in Section~\ref{subsec:baseline}. Next, Genetic Algorithm (GA) is also considered as another baseline approach in Section~\ref{subsec:GA}. Finally, the comparative evaluation of all the identified neural architectures is discussed in Section~\ref{subsec:comparative_evaluation}.
\end{itemize}

\subsection{Experimental Setup}
\label{subsec:search}

Since the simultaneous evolution of feature selection and neural architecture is posed as a generic multi-objective optimization problem (see Section~\ref{sec:NAS}), any suitable search algorithm can be used for the neural architecture search. Accordingly, this study, also considers Genetic Algorithm (GA) as one of the baseline approach. GA has been used to address various aspects of neural architecture search in several prior studies, which include but are not limited to number of hidden neurons and layers~\citep{Stathakis:2009}. It is, therefore, pertinent to benchmark the search performance of 2DS in relation to GA.

In this study, GA is configured as follows: \textit{binary-tournament selection, uniform crossover} and \textit{bit-flip mutation}. The selection of uniform crossover is, in part, inspired form the earlier investigation of authors~\citep{Hafiz:Swain:CEC:2018,Hafiz:Swain:MOEA:2020}, which suggest that \textit{single-point} crossover operator of the simple GA is not likely to perform well for the combinatorial problems similar to feature selection. For such problems, the desired \textit{schemas} may not be of low order, hence, the underlying assumptions of simple GA may not hold, see~\citep{Spears:DeJong:1995,Mitchell:1998,Hafiz:Swain:MOEA:2020} for a detailed treatment on this issue.

The parameters of search algorithms are fixed as follows: GA - population size ($30$), crossover probability ($0.8$) and mutation probability ($\frac{1}{n}$); and 2DS - population size ($ps = 30$), unification factor ($u_f = 0.9$), refresh gap ($RG=30$). A candidate solution ($\mathcal{X}$) in both GA and 2DS represents a combination of neural architecture and feature subset, as discussed in Section~\ref{sec:NAS}. For search purposes, each candidate solution is encoded using the binary string representation given in Section~\ref{subsec:encoding}. The performance of each candidate solution under consideration is determined following the steps outlined in Algorithm~\ref{al:crf}. To accommodate the stochastic nature of GA and 2DS, multiple independent runs are being carried out. Each run is set to terminate after a total of $6000$ evaluations of the criterion function, $J(\cdot)$. The best solution identified over such 20 independent runs is considered for the further analysis.

Finally, the overall NASDAQ time-series data is used as follows:
\begin{itemize}
    \smallskip
    \item \textbf{Training}: depending on the \textit{learning scenario} only \textit{within-COVID data} ($\mathcal{D}^{cv}$) or the combination of \textit{pre-} and \textit{within-COVID data} ($\mathcal{D}^{pr} \cup \mathcal{D}^{cv}$) is used for the weight estimation (see Section~\ref{subsec:multidataset_learning}, Algorithm~\ref{algo:eff_LS_F} and~\ref{algo:eff_LS_eps}).
    \smallskip
    \item \textbf{Testing during Search}: During the search process the efficacy of the candidate neural architectures is being evaluated by using the \textit{test-data} ($\mathcal{D}_{test}$).
    \smallskip
    \item \textbf{Generalization Test}: Once a suitable neural architecture is identified by the search algorithm, its generalization capability is evaluated using the \textit{hold-out data} ($\mathcal{D}_{hold}$). Note that this dataset is purposefully kept aside and not being used in any step of the search process to get the unbiased estimate of the generalization capability. 
    \smallskip
\end{itemize}

\subsection{Comparative Evaluation of Learning Scenarios: Influence of Data Incompatibility}
\label{subsec:comparison_of_learning_scenarios}

The first part of this investigation test the hypothesis that distinct market behavior in pre- and within-COVID time window may hinder the learning process in neural networks. To this end, neural architecture are evolved to predict day-ahead NASDAQ index movement under both \textit{full} ($\mathcal{LS_F}$) and \textit{split} ($\mathcal{LS}_\varepsilon$) learning scenarios (see Section~\ref{subsec:multidataset_learning}). 

The efficacy of candidate neural architectures in the \textit{full} and \textit{split} learning scenario is determined following Algorithm~\ref{algo:eff_LS_F} and~\ref{algo:eff_LS_eps}, respectively. The remaining search environment (see Section~\ref{subsec:search}) is unchanged under both the learning scenarios. In both the learning scenarios, the neural architectures are identified for different set of \textit{preference weights} using 2DS. Table~\ref{t:bestl2l3} gives the \textit{best} neural architecture identified over 20 independent runs of 2DS for different combinations of \textit{preferences} ($\Theta$) and the learning scenarios. The complexity $\mathcal{C}$ of these architectures is determined using~(\ref{eq:complexity}) and it is also reported along with the classification accuracy and MCC over the \textit{hold-out} dataset ($\mathcal{D}_{hold}$) in Table~\ref{t:bestl2l3}. 

The results in Table~\ref{t:bestl2l3} clearly indicate that the \textit{split} learning scenario ($\mathcal{LS}_\varepsilon$) is likely to evolve better neural architectures irrespective of the \textit{preference} of the decision maker. To further examine the statistical significance of these outcomes, Wilcoxon signed rank sum test is applied to evaluate the following \textit{null} and \textit{alternative} hypotheses:
\begin{itemize}
    \smallskip
    \item[] Null Hypothesis ($H_0$): $\mu_\varepsilon = \mu_\mathcal{F}$, \qquad \qquad Alternate Hypothesis ($H_1$) : $\mu_\varepsilon > \mu_\mathcal{F}$
    \smallskip
\end{itemize}
where, $\mu_\mathcal{F}$ and $\mu_\varepsilon$ represent average accuracy over $\mathcal{D}_{hold}$ corresponding to neural architectures identified over multiple runs under $\mathcal{LS_F}$ and $\mathcal{LS}_\varepsilon$, respectively. The similar procedure is repeated for MCC outcomes.

Table~\ref{t:statl2l3} gives the average and variance of the \textit{hold-out} accuracy and MCC across different neural architectures which are identified under a particular learning scenario. The $p-$value of null hypothesis following the Wilcoxon signed rank sum test is also shown. It is clear that the \textit{null} hypothesis can easily be rejected at $\alpha=0.05$ \textit{significance} level for both accuracy and MCC.

Given the \textit{significant} performance improvement of $\mathcal{LS}_\varepsilon$ over $\mathcal{LS_F}$, it is safe to infer that the distinct pre-COVID behavior does adversely affect the estimation of network weights. Further, these results give the empirical proof to the fact that it is possible to reconcile the the underlying concordant information in the pre-COVID time-window through the $\epsilon-$constraint framework of $\mathcal{LS}_\varepsilon$.

Following these outcomes, the neural architecture search is investigated under the \textit{split} learning scenario, $\mathcal{LS}_\varepsilon$, for the reminder of this study.


\begin{table*}[!t]
  \centering
  \caption{Best neural topologies evolved out of 20 independent runs of 2DS under different learning scenarios}
  \label{t:bestl2l3}%
  \begin{adjustbox}{max width=\textwidth}
  \small
  \begin{threeparttable}

    \begin{tabular}{cccccccccc}
    \toprule
    \multirow{2}[4]{*}{\makecell{\textbf{Preference}\\\textbf{Weights$^\dagger$}}} & \multirow{2}[4]{*}{\makecell{\textbf{Learning}\\\textbf{Scenario}}} &  \multirow{2}[4]{*}{\makecell{\textbf{Selected}\\\textbf{Features,}\\ \boldmath{$\#X$}}} & \multicolumn{2}{c}{\textbf{Hidden Layer -1}} & \multicolumn{2}{c}{\textbf{Hidden Layer -2}} & \multirow{2}[4]{*}{\makecell{\textbf{Complexity,}\\\boldmath{$\mathcal{C}$}}} & \multirow{2}[4]{*}{\makecell{\textbf{Hold-out}\\\textbf{Accuracy}\\($\%$)}} & \multirow{2}[4]{*}{\makecell{\textbf{Hold-out}\\\textbf{MCC}}} \\[1ex]
    \cmidrule{4-7}          &       &       & \makecell{\textbf{Size,}\\\boldmath $s^1$} & \makecell{\textbf{Activation}\\\textbf{Function,}\\\boldmath$f^1$} & \makecell{\textbf{Size,}\\\boldmath $s^2$} & \makecell{\textbf{Activation}\\\textbf{Function,}\\\boldmath $f^2$} &       &       &  \\[1ex]
    \midrule
    \multirow{2}{*}{$\Theta_1$} & $\mathcal{LS_F}$ & 5     & 64    & \textit{tansig} & - & - & 0.3592 & 53.02 & 0.0718 \\[0.5ex]
    & $\mathcal{LS}_\varepsilon$ & 4 & 64 & \textit{tansig} & - & - & 0.3543 & \textbf{55.11} & \textbf{0.1127} \\[0.5ex]
    \midrule
    \multirow{2}{*}{$\Theta_2$} & $\mathcal{LS_F}$ & 10    & 98    & \textit{tansig} & 32    & \textit{tansig} & 0.5530 & 51.76 & 0.0412 \\[0.5ex]
    & $\mathcal{LS}_\varepsilon$ & 4     & 16    & \textit{tansig} & -     & -     & 0.2283 & \textbf{57.09} & \textbf{0.1598} \\[0.5ex]
    \midrule
    \multirow{2}{*}{$\Theta_3$} & $\mathcal{LS_F}$ & 9     & 12    & \textit{tansig} & - & - & 0.2423 & 52.20 & 0.0487 \\[0.5ex]
    & $\mathcal{LS}_\varepsilon$ & 9     & 64    & \textit{tansig} & - & - & 0.3788 & \textbf{56.26} & \textbf{0.1522} \\[0.5ex]
    \bottomrule

    \end{tabular}%
    
    \begin{tablenotes}
      \footnotesize
      \item $\dagger$ $\Theta_1 = [ \theta_\mathcal{E}, \theta_\mathcal{C}] = [0.75, 0.25]$; $\Theta_2 = [0.5, 0.5]$ and $\Theta_3 = [0.25, 0.75]$
    \end{tablenotes}
    
  \end{threeparttable}
 \end{adjustbox}
\end{table*}%
\begin{table*}[!t]
  \centering
  \caption{The null hypothesis testing for similarity of neural architectures identified under $\mathcal{LS_F}$ and  $\mathcal{LS}_\epsilon$ scenarios$^\dagger$}
  \label{t:statl2l3}%
  \small
  \begin{adjustbox}{max width=0.85\textwidth}
  \begin{threeparttable}

    \begin{tabular}{cccccccc}
    \toprule
    \multirow{2}[4]{*}{\makecell{\textbf{Learning}\\\textbf{Scenario}}} &  & \multicolumn{2}{c}{\boldmath{$\Theta_1$}} & \multicolumn{2}{c}{\boldmath{$\Theta_2$}} & \multicolumn{2}{c}{\boldmath{$\Theta_3$}} \\[1ex]
    \cmidrule{3-8} &  & \makecell{\textbf{Topological}\\\textbf{Accuracy}} & \makecell{\textbf{Topological}\\\textbf{MCC}} & \makecell{\textbf{Topological}\\\textbf{Accuracy}} & \makecell{\textbf{Topological}\\\textbf{MCC}} & \makecell{\textbf{Topological}\\\textbf{Accuracy}} & \makecell{\textbf{Topological}\\\textbf{MCC}} \\[1ex]
    \midrule
    \multirow{2}[2]{*}{$\mathcal{LS}_F$} & Mean & 50.46 & 0.0047 & 50.08 & -0.0040 & 50.27 & -0.0026 \\[0.5ex]
          & Variance & $1.5\times 10^{-2}$ & $3.9\times 10^{-2}$ & $7.11\times 10^{-3}$ & $2.08\times 10^{-2}$ & $1.02\times 10^{-2}$ & $2.90\times 10^{-2}$ \\[0.5ex]
    \midrule
    \multirow{2}[2]{*}{$\mathcal{LS}_\varepsilon$}& Mean &  51.66 & 0.0331 & 51.44 & 0.0240 & 52.12 & 0.0430 \\[0.5ex]
          & Variance & $1.6\times 10^{-2}$ & $3.9\times 10^{-2}$ & $1.94\times 10^{-2}$ & $4.81\times 10^{-2}$ & $2.19\times 10^{-2}$ & $5.65\times 10^{-2}$ \\[0.5ex]
    \midrule
    \multicolumn{2}{c}{\makecell{$p-$value$^\ddagger$\\Null Hypothesis ($H_0$)}} & \makecell{$1.38\times 10^{-2}$\\ \xmark} & \makecell{$1.76\times 10^{-2}$\\ \xmark} & \makecell{$5.90\times 10^{-3}$\\ \xmark} & \makecell{$1.51\times 10^{-2}$\\ \xmark} & \makecell{$2.20\times 10^{-3}$\\ \xmark} & \makecell{$4.40\times 10^{-3}$\\ \xmark} \\
    \bottomrule
    \end{tabular}%

    \begin{tablenotes}
      \footnotesize
      \item $\dagger$ $\Theta_1 = [ \theta_\mathcal{E}, \theta_\mathcal{C}] = [0.75, 0.25]$; $\Theta_2 = [0.5, 0.5]$ and $\Theta_3 = [0.25, 0.75]$
      
      \item $\ddagger$ $p-$value of Wilcoxon signed rank sum test; Null-hypothesis at $95\%$ confidence interval
    \end{tablenotes}
  \end{threeparttable}
 \end{adjustbox}
\end{table*}%
\subsection{Dynamic Search Behavior of 2DS}
\label{subsec:2DS_convergence}

\begin{figure*}[!t]
\centering

\begin{subfigure}{.31\textwidth}
  \centering
  \includegraphics[width=\textwidth]{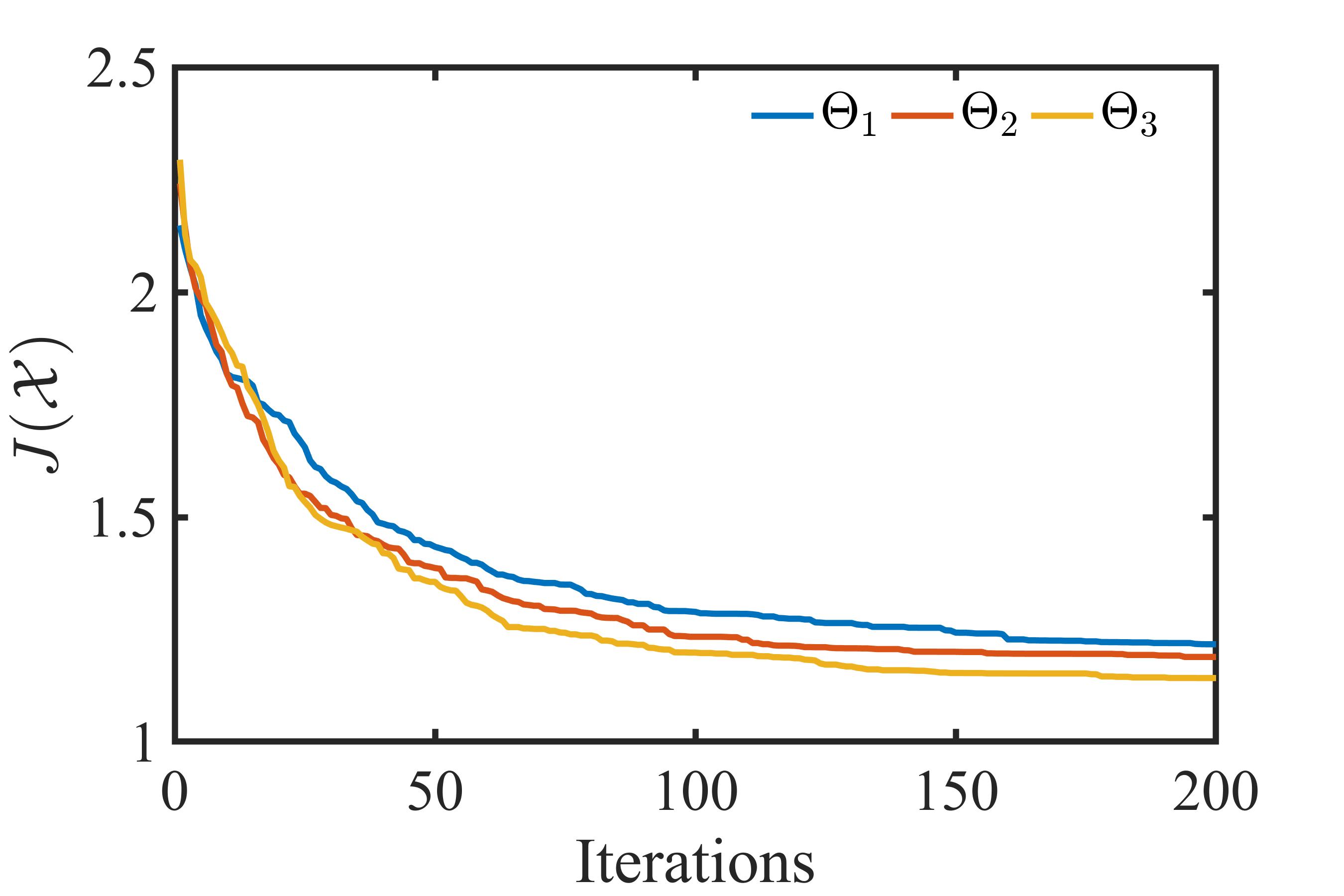}
  \caption{criterion function}
  \label{f:2ds_j}
\end{subfigure}
\hfill
\begin{subfigure}{.31\textwidth}
  \centering
  \includegraphics[width=\textwidth]{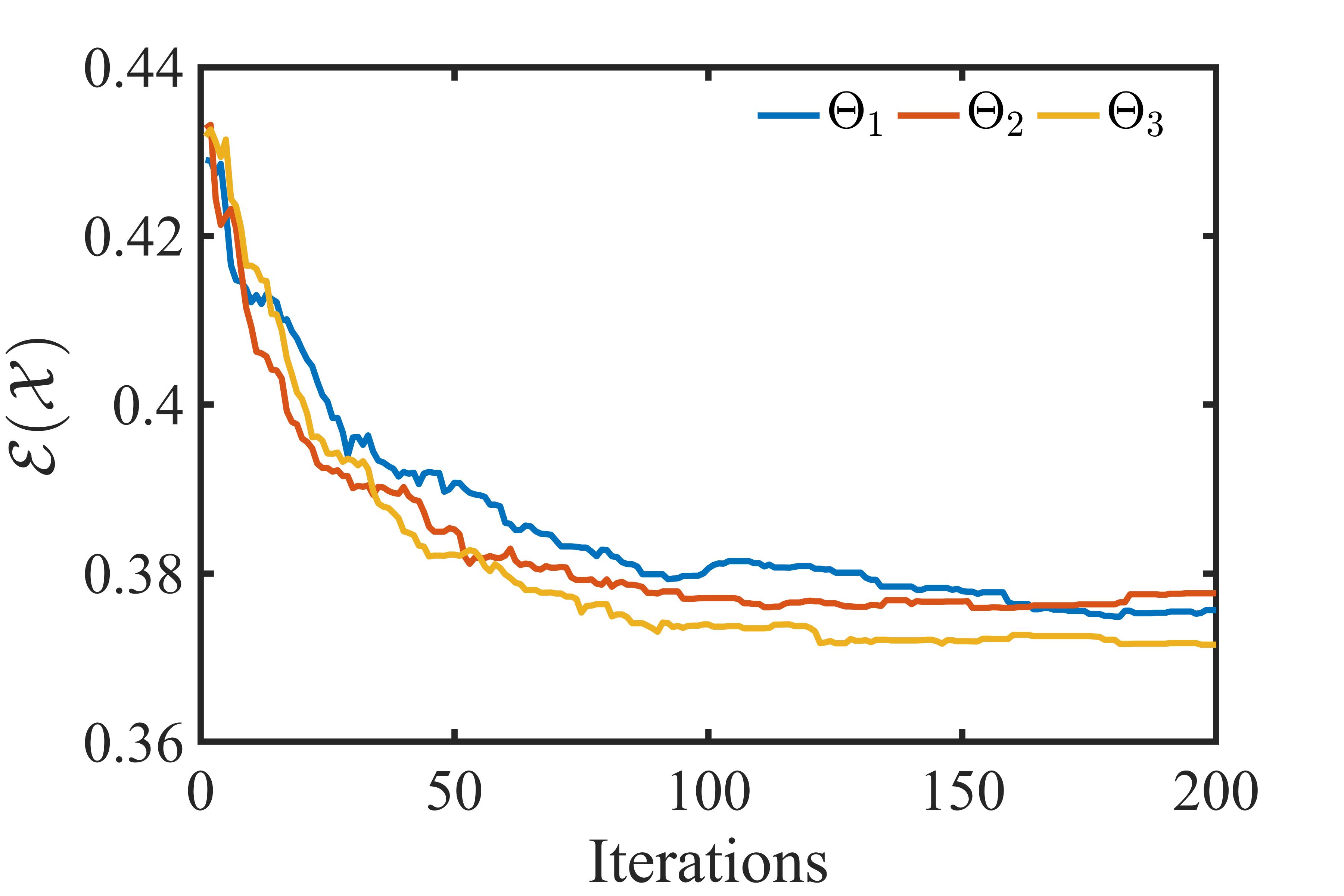}
  \caption{classification error over test data, $\mathcal{D}_{test}$}
  \label{f:2ds_e}
\end{subfigure}
\hfill
\begin{subfigure}{.31\textwidth}
  \centering
  \includegraphics[width=\textwidth]{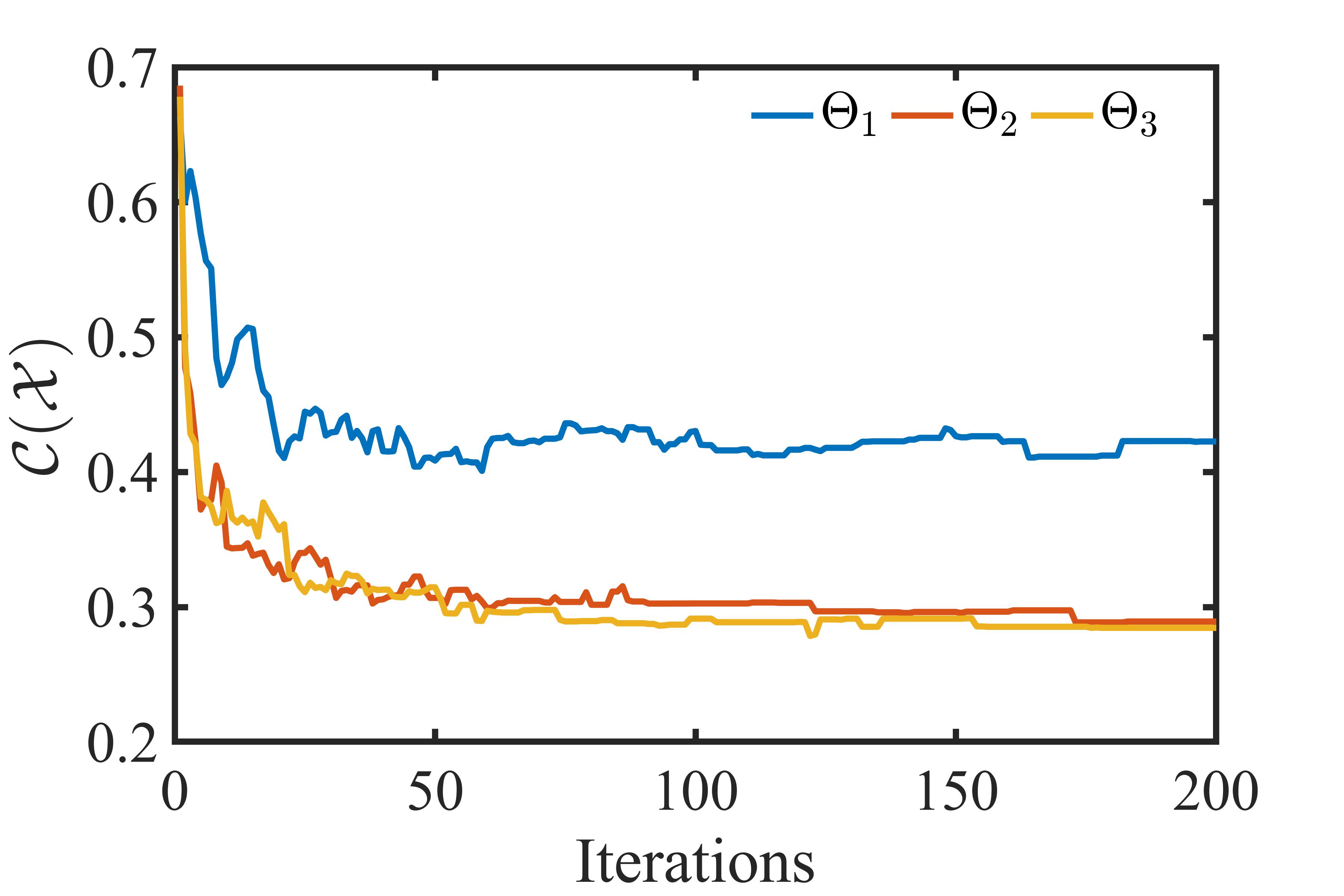}
  \caption{overall architectural complexity}
  \label{f:2ds_c}
\end{subfigure}
\medskip
\begin{subfigure}{.31\textwidth}
  \centering
  \includegraphics[width=\textwidth]{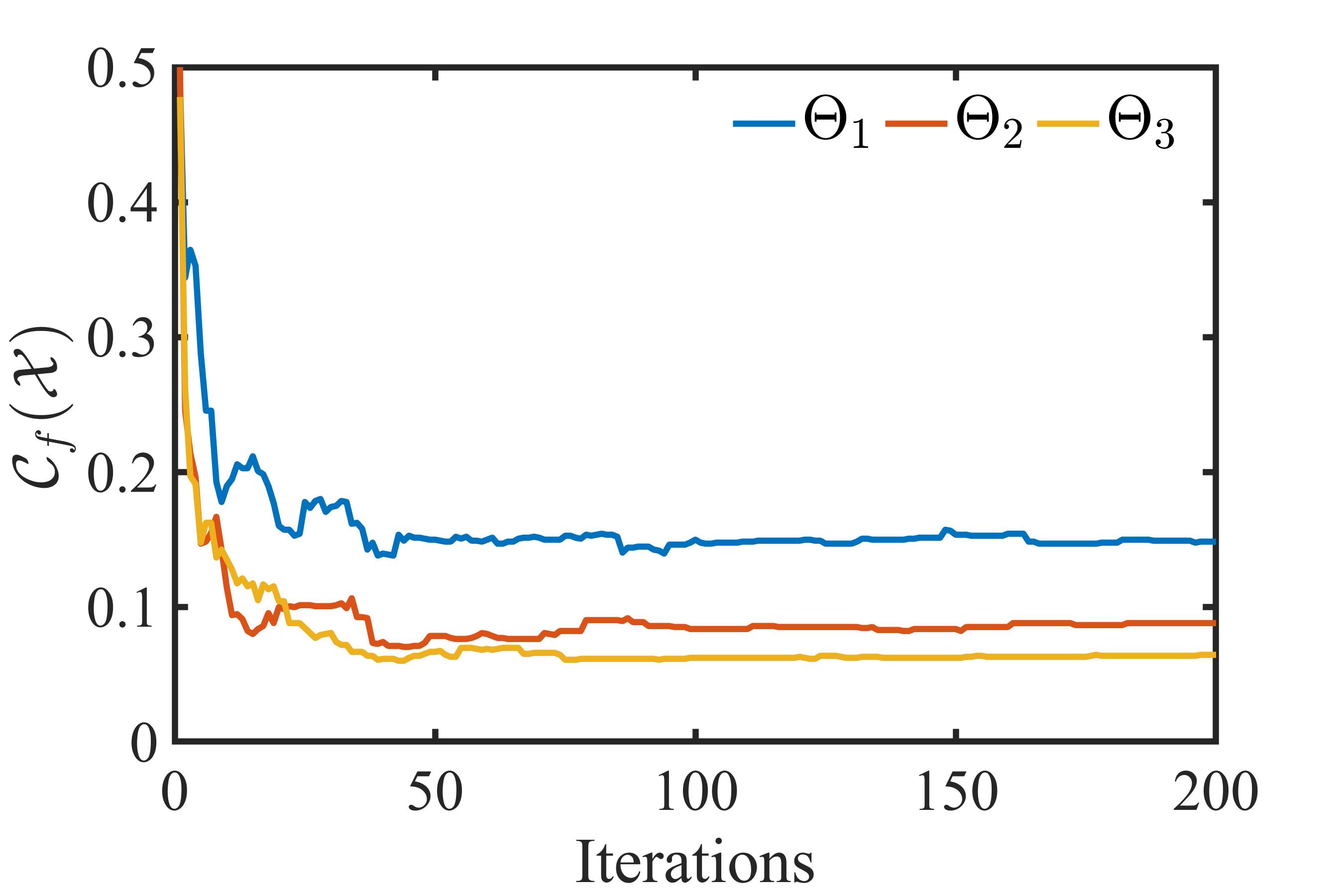}
  \caption{feature complexity}
  \label{f:2ds_cf}
\end{subfigure}
\hfill
\begin{subfigure}{.31\textwidth}
  \centering
  \includegraphics[width=\textwidth]{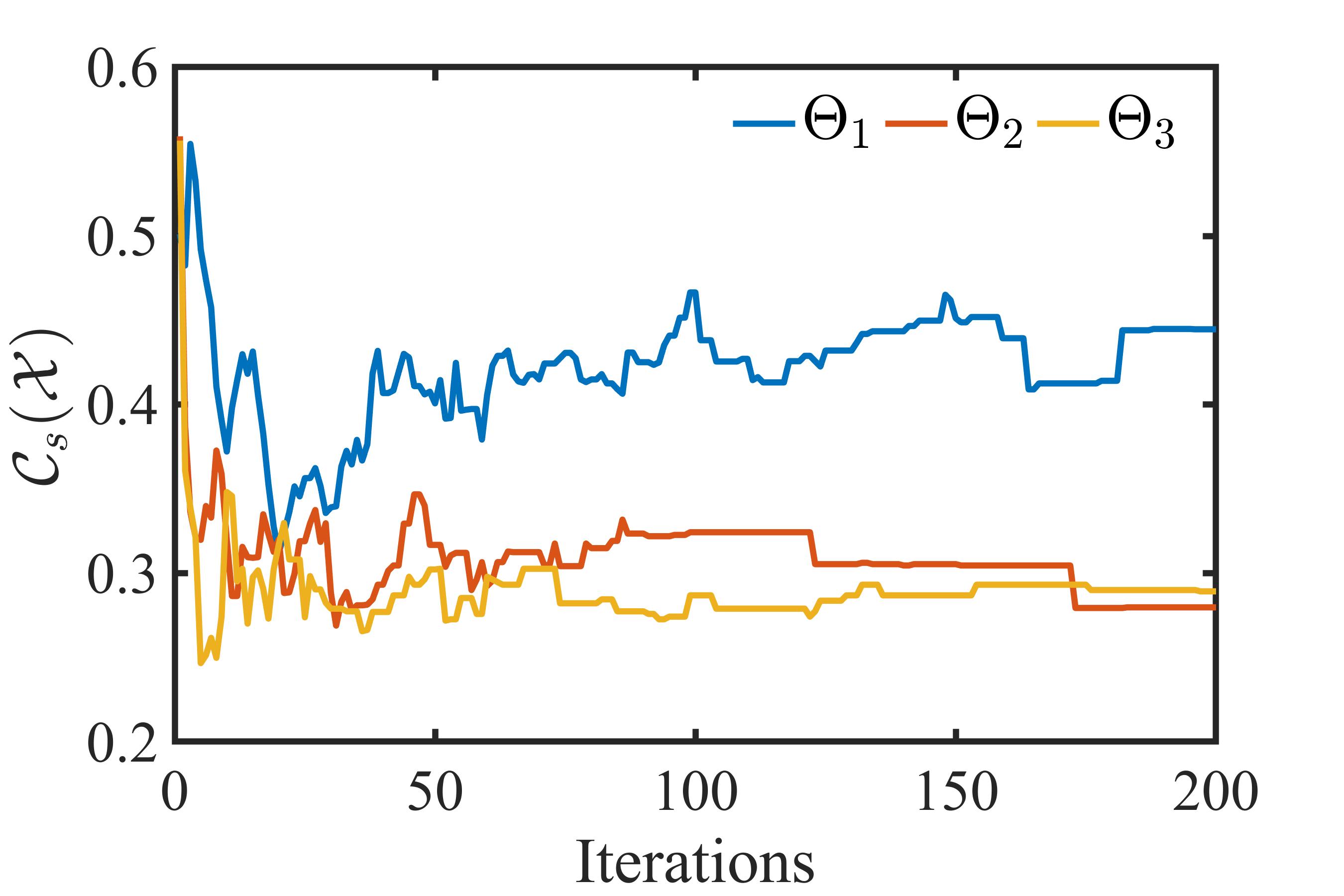}
  \caption{neuron complexity}
  \label{f:2ds_cs}
\end{subfigure}
\hfill
\begin{subfigure}{.31\textwidth}
  \centering
  \includegraphics[width=\textwidth]{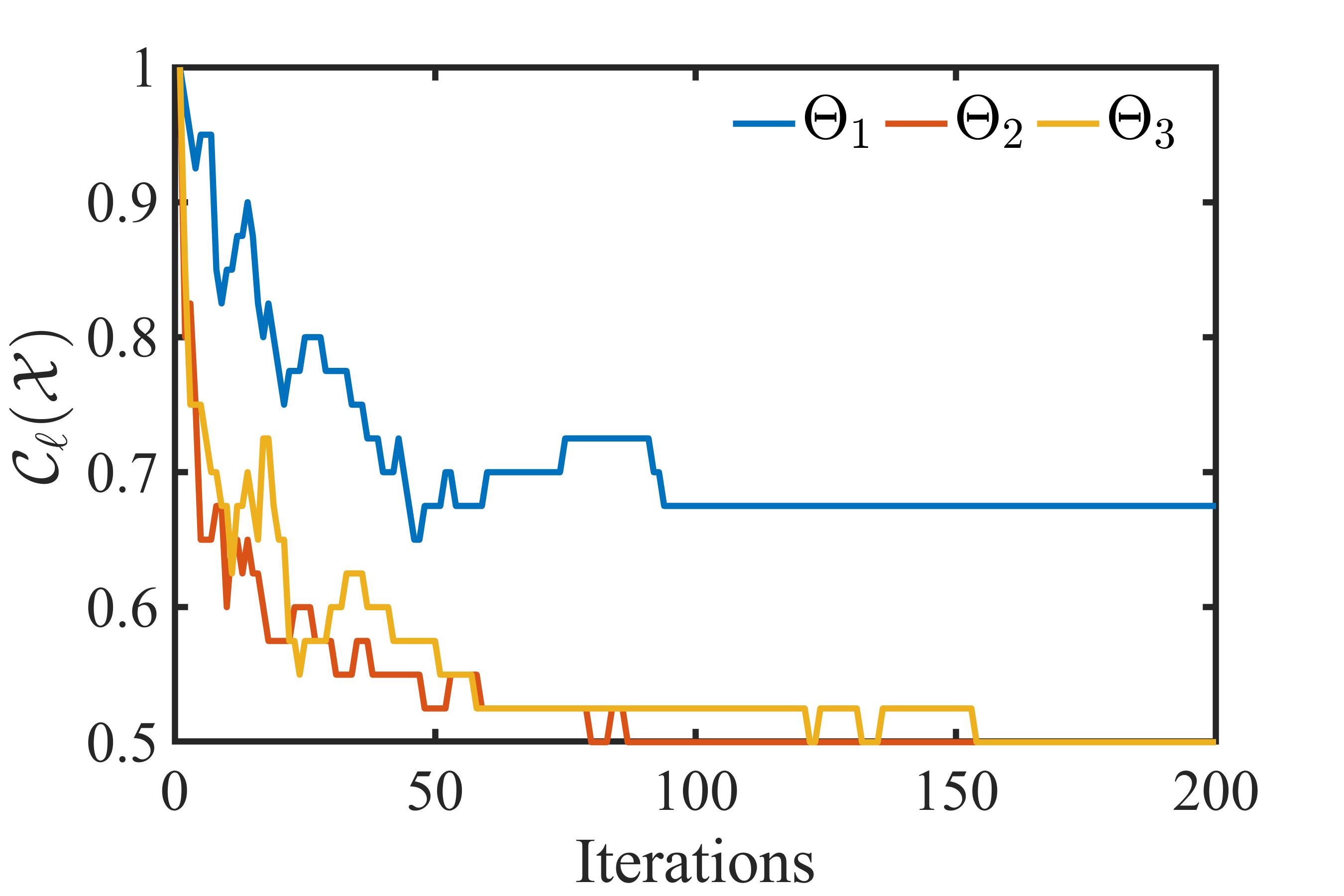}
  \caption{layer complexity}
  \label{f:2ds_cl}
\end{subfigure}

\caption{Convergence plots averaged over $20$ runs of 2DS in the \textit{split-learning} scenario, $\mathcal{LS}_\varepsilon$.}
\label{f:conv2DS}
\end{figure*}

To gain insight into the search process, the convergence behavior of 2DS was recorded and reproduced here in Fig~\ref{f:conv2DS}. As expected, there is a simultaneous reduction in the classification error (Fig.~\ref{f:2ds_e}) and architectural complexity (Fig.~\ref{f:2ds_c}). These observations give further merit to the hypothesis that the \textit{Occam's razor} or \textit{principle-of-parsimony} holds for the neural design, at least for the problem being considered here. 

Further, the implications of the decision maker's (DM) preferences are clearly visible; the complexity is the highest with $\Theta_1$ (see Fig.~\ref{f:2ds_c}) where \textit{complexity} is given a lower priority, \textit{i.e.}, $\Theta_1 = [\theta_\mathcal{E}, \ \theta_\mathcal{C}] = [0.75, \ 0.25]$. To analyze this further, the architecture complexity is decomposed and observed as follows:
\begin{small}
\begin{align}
   \label{eq:complexity}
   \mathcal{C}(\mathcal{X}_i) & = \frac{1}{3} \Big\{ \mathcal{C}_f(\mathcal{X}_i) + \mathcal{C}_s(\mathcal{X}_i) + \mathcal{C}_\ell(\mathcal{X}_i) \Big\}\\
   \text{where,} \quad \mathcal{C}_f(\mathcal{X}_i) & = \frac{\# (X_i)}{n_f}, \quad \mathcal{C}_s(\mathcal{X}_i) = \sum \limits_{k=1}^{n_\ell} \frac{s^k}{s^{max}}, \quad \mathcal{C}_s(\mathcal{X}_i) = \frac{\# \Big( \begin{Bmatrix} (s^k,f^k) \ \big| \ s^k \neq 0, & \forall k \in [1,n_\ell] \end{Bmatrix} \Big)}{n_\ell}\nonumber
\end{align}
\end{small}
$\mathcal{C}_f(\cdot)$, $\mathcal{C}_\ell(\cdot)$ and $\mathcal{C}_s(\cdot)$ respectively denote the \textit{complexity} of \textit{feature}, \textit{layer} and \textit{neurons}; $X_i$ gives the feature subset embedded within $\mathcal{X}_i$; $s^{max}$ and $n_\ell$ respectively give maximum number of hidden neurons and hidden layers; $n_f$ denotes the total number of features; and $\#(\cdot)$ determines the \textit{cardinality} of a particular set.

The \textit{constituent} component of the architectural complexity are shown in Fig.~\ref{f:2ds_cf} (\textit{feature complexity}, $\mathcal{C}_f$), Fig.~\ref{f:2ds_cs} (\textit{neuron complexity}, $\mathcal{C}_s$) and Fig.~\ref{f:2ds_cf} (\textit{layer complexity}, $\mathcal{C}_\ell$). The results show that, irrespective of the \textit{preference} specification, 2DS selects around only $8\%-18\%$ of the total features, see Fig.~\ref{f:2ds_cf}. Further, the layer complexity (Fig.~\ref{f:2ds_cl}) reduces to $\mathcal{C}_\ell=0.5$ (with $\Theta_2$ and $\Theta_3$) and to $\mathcal{C}_\ell=0.7$ (with $\Theta_1$) which implies that in most runs only a single hidden layer is being selected. Similar behavior is observed in the \textit{neuron complexity} (Fig.~\ref{f:2ds_cs}); approximately $30\%$ of total neurons are being used with $\Theta_2$ and $\Theta_3$ whereas this increases to around $45\%$ with $\Theta_1$.

Given that there is no significant trade-off in classification error, we recommend either the \textit{balanced} ($\Theta_2$) or \textit{sparse} ($\Theta_3$) selection of preference weights, \textit{i.e.}, $\Theta_2 = [\theta_\mathcal{E}, \ \theta_\mathcal{C}] = [0.5, \ 0.5]$ or $\Theta_3 = [\theta_\mathcal{E}, \ \theta_\mathcal{C}] = [0.25, \ 0.75]$.

\subsection{Baseline Approach: Empirical Rules for Neural Design}
\label{subsec:baseline}

This part of the investigation focuses on empirical design rules for neural network design, which will serve as the \textit{baseline} approach for the comparative evaluation purposes. The objective here is to highlight the need for neural architecture search. 

While there is no consensus on the neural architecture design for stock forecasting, in many cases~\citep{Yu:Chen:2008}, a single hidden layer neural network is selected wherein the size of hidden layer is set to two times that of input neurons, \textit{i.e.}, $s = 2 \times n_f$. The roots of this rule-of-thumb can be traced back to the extension of Kolmogorov's theorem to neural design in~\citep{Hecht-Nielsen:1987}. In addition to the Kolmogorov's theorem, this study considers five other  empirical rules for neural design, which are shown in~\ref{sec:app}. The common rationale behind these rules is to tie the network complexity to number of input features, output classes and often to the number of training samples, as seen in~\ref{sec:app}. A detailed treatment on the empirical neural design rules can be found in~\citep{Stathakis:2009} and the references therein.

It is worth noting that while the empirical design rules provide guidelines for the selection of hidden neurons, they do not address the selection of activation function.
Further, the issue of feature selection is also not considered. Given that appropriate selection of both, features and activation function, is likely influence the classification performance, the following two extensions are proposed two the empirical design rules:
\smallskip
\begin{itemize}
    \item \textit{Feature Selection}: for sake of of fair comparison, a \textit{filter} type feature selection algorithm, mRmR~\citep{Peng:Long:2005}, is being applied to remove redundant and irrelevant features in the baseline approach. The empirical rules are applied to both, full feature set and the reduced feature set identified using mRmR. It is worth noting that the evaluation of mutual information in mRmR requires that the features are discretized beforehand. For this purpose, the features were discretized using CAIM algorithm~\citep{Kurgan:Cios:2004}. Further, mRmR requires an \textit{apriori} specification of the cardinality of the reduced subset (number of features). In this study, cardinality of the reduced subset was varied in the range of $[0.1\times n_f, 0.75\times n_f]$. The results of this empirical investigation suggested that the cardinality of the reduced subset in mRmR should be set to $n_f=31$. Note that the results pertaining to this part of investigation are omitted here due to space constraints.
    
    \smallskip
    
    \item \textit{Activation Function}: Both `\textit{sigmoid}' and `\textit{tangent sigmoid}' activation functions were considered with each empirical rule. An improved classification performance was obtained with the \textit{tangent sigmoid}, hence, only the results obtained with this activation function are discussed in the following.
    
    \smallskip
\end{itemize}

\begin{table*}[!t]
  \centering
  \caption{Performance of the neural architectures determined following the empirical rules in~\ref{sec:app}$^\dagger$}
  \label{t:benchmark}%
  \begin{adjustbox}{max width=\textwidth}
  \scriptsize
  \begin{threeparttable}

    \begin{tabular}{ccccccc}
    \toprule
    \textbf{Scenario} & \makecell{\textbf{Topology}\\ \textbf{Rule}} & \makecell{\textbf{Size of}\\\textbf{Layer - 1}\\\boldmath$(s^1)$} & \makecell{\textbf{Size of}\\\textbf{Layer - 2}\\\boldmath$(s^2)$} & \makecell{\textbf{Complexity}\\\boldmath$\mathcal{C(\cdot)}$} & \makecell{\textbf{Hold-out}\\\textbf{Accuracy}\\\boldmath{$\eta(\mathcal{X}_i,\mathcal{D}_{hold})$}} & 
    \makecell{\textbf{Hold-out}\\\textbf{MCC}\\\boldmath{$\Phi(\mathcal{X}_i,\mathcal{D}_{hold})$}}\\[1ex]
    \midrule
    
    \multirow{6}{*}{\makecell{All Features\\$(n_f = 68)$}} & Kolmogorov & 137   & -     & 0.5894 & 51.43 & 0.0291 \\[0.5ex]
    & Hush & 272   & -     & 0.6774 & 51.04 & 0.0164 \\[0.5ex]
    & Wang & 45    & -     & 0.5294 & 51.15 & 0.0156 \\[0.5ex]
    & Ripley & 35    & -     & 0.5228 & 51.32 & 0.0274 \\[0.5ex]
    & \textbf{Fletcher-Goss} & \textbf{18}    & -     & \textbf{0.5117} & \textbf{51.70} & \textbf{0.0376} \\[0.5ex]
    & Huang & 57    & 19    & 0.6915 &  51.32 & 0.0204 \\[0.5ex]
    \midrule
    \multirow{6}{*}{\makecell{mRmR\\$(n_f = 31)$}} & Kolmogorov & 63    & -     & 0.3597 & 52.31 & 0.0615 \\[0.5ex]
    & Hush & 124   & -     & 0.3995 & 51.37 & 0.0263 \\[0.5ex]
    & Wang & 21    & -     & 0.3323 & 52.53 & 0.0598 \\[0.5ex]
    
    & Ripley & 17    & -     & 0.3297 & 51.70 & 0.0368 \\[0.5ex]
    & \textbf{Fletcher-Goss} & \textbf{13}    & -     & \textbf{0.3271} & \textbf{52.69} & \textbf{0.0686} \\[0.5ex]
    & Huang & 57    & 19    & 0.5101 & 52.69 & 0.0573 \\
    \bottomrule
    \end{tabular}%

    \begin{tablenotes}
      \scriptsize
      \item $\dagger$ `$tansig$' is selected as the activation function for hidden layer/s
      
    \end{tablenotes}
  \end{threeparttable}
 \end{adjustbox}
\end{table*}%

To emulate the mixed learning scenario, $\mathcal{LS}_\varepsilon$, mRmR based feature selection was carried out using the combination of COVID ($\mathcal{D}^{cv}$) and Pre-COVID data ($\mathcal{D}^{pr}$). Once the reduced feature subset is found, the neural architecture corresponding to each empirical rule is determined and the weights of the subsequent network are trained using the COVID training data, $\mathcal{D}^{cv}$. Finally, the performance of the trained network is determined using the COVID test ($\mathcal{D}_{test}$) and hold-out datasets ($\mathcal{D}_{hold}$). The efficacy of each neural architecture is determined by considering an average performance over $20$ learning cycles as discussed in Section~\ref{subsec:NeuEff} and outlined in the following: 
\begin{small}
\begin{align}
    \eta(\mathcal{X}_i,\mathcal{D}_{hold}) & = \displaystyle\frac{100}{cycles} \sum \limits_{k=1}^{cycles} 1 - \mathcal{E}_k(\mathcal{X}_i,\mathcal{W}_{i,k}^{\ast},\mathcal{D}_{hold}), \quad \textit{where,} \ 
   \mathcal{W}_{i,k}^{\ast} = \argmin \limits_{\mathcal{W}} \ {\rm L}(\mathcal{X}_i,\mathcal{W},\mathcal{D}^{cv})\\
  \Phi(\mathcal{X}_i,\mathcal{D}_{hold}) & =  \displaystyle \frac{1}{cycles} \sum \limits_{k=1}^{cycles} \Phi_k(\mathcal{X}_i,\mathcal{W}_{i,k}^{\ast},\mathcal{D}_{hold}) \nonumber
 \end{align}
\end{small} 
$\eta(\mathcal{X}_i,\mathcal{D}_{hold})$ and $\Phi(\mathcal{X}_i,\mathcal{D}_{hold})$ respectively give the percentage classification accuracy and MCC of the $i^{th}$ neural architecture $\mathcal{X}_i$ over the hold-out dataset, $\mathcal{D}_{hold}$; and $cycles=20$.

Table~\ref{t:benchmark} gives the classification performance of neural networks designed following these steps. The combined sparsity of feature subset and neural architecture is also determined using~(\ref{eq:complexity}) and listed in Table~\ref{t:benchmark} for comparison purposes. The results indicate that, as expected, the feature selection in general has a positive effect on the classification performance. The neural architectures with reduced feature subset have relatively better MCC and Accuracy on the hold-out dataset. Further, the Fletcher-Goss's (FG) rule gives the better neural architecture for both scenarios, full and reduced feature set. It is interesting to see that among the compared empirical rules, the least complex architecture is obtained following the FG rule, as seen in Table~\ref{t:benchmark}. For the remainder of this study, the combination of FG rule and Full feature set is referred to as `\textit{Full}' architecture; similarly `\textit{Reduced}' denotes the neural architecture derived from the mRmR selected features and FG rule. These two architectures are shown in \textit{bold-face} in Table~\ref{t:benchmark} and are considered for the further comparative evaluation.

It is clear that the performance of neural networks designed using any combination of baseline approach is below the usual acceptable threshold of $55\%$ classification accuracy~\citep{UlHaq2021}. These outcomes clearly highlight the scope for further improvement in the neural design.

\subsection{Baseline Approach: Neural Architecture Search with GA}
\label{subsec:GA}

To benchmark the search performance of 2DS, GA is also applied to identify the optimum neural architecture for the prediction of NASDAQ movements. To this end, 20 independent runs of GA are carried out following the search environment outlined in Section~\ref{subsec:search}, under the \textit{split-dataset} learning scenario, $\mathcal{LS}_\varepsilon$. For this test, \textit{balanced} preference specifications are considered, \textit{i.e.}, $\Theta_2 = [\theta_\mathcal{E}, \ \theta_\mathcal{C}] = [0.5, \ 0.5]$.

The \textit{best} neural architecture which is identified from $20$ independent runs of GA is given in Table~\ref{t:resGA}. It is clear that this architecture is relatively more complex ($\mathcal{C}=0.6376$, see Table~\ref{t:resGA}) compared to the neural architecture identified by 2DS under the similar environment($\mathcal{C}=0.2283$, see Table~\ref{t:bestl2l3}). This increase in the architectural complexity with GA does not, however, translate into a better classification performance; the accuracy and MCC over the $\mathcal{D}_{hold}$ is limited to $52\%$ and $0.0381$ (see Table~\ref{t:resGA}).

\begin{figure*}[!t]
\centering

\begin{subfigure}{.245\textwidth}
  \centering
  \includegraphics[width=\textwidth]{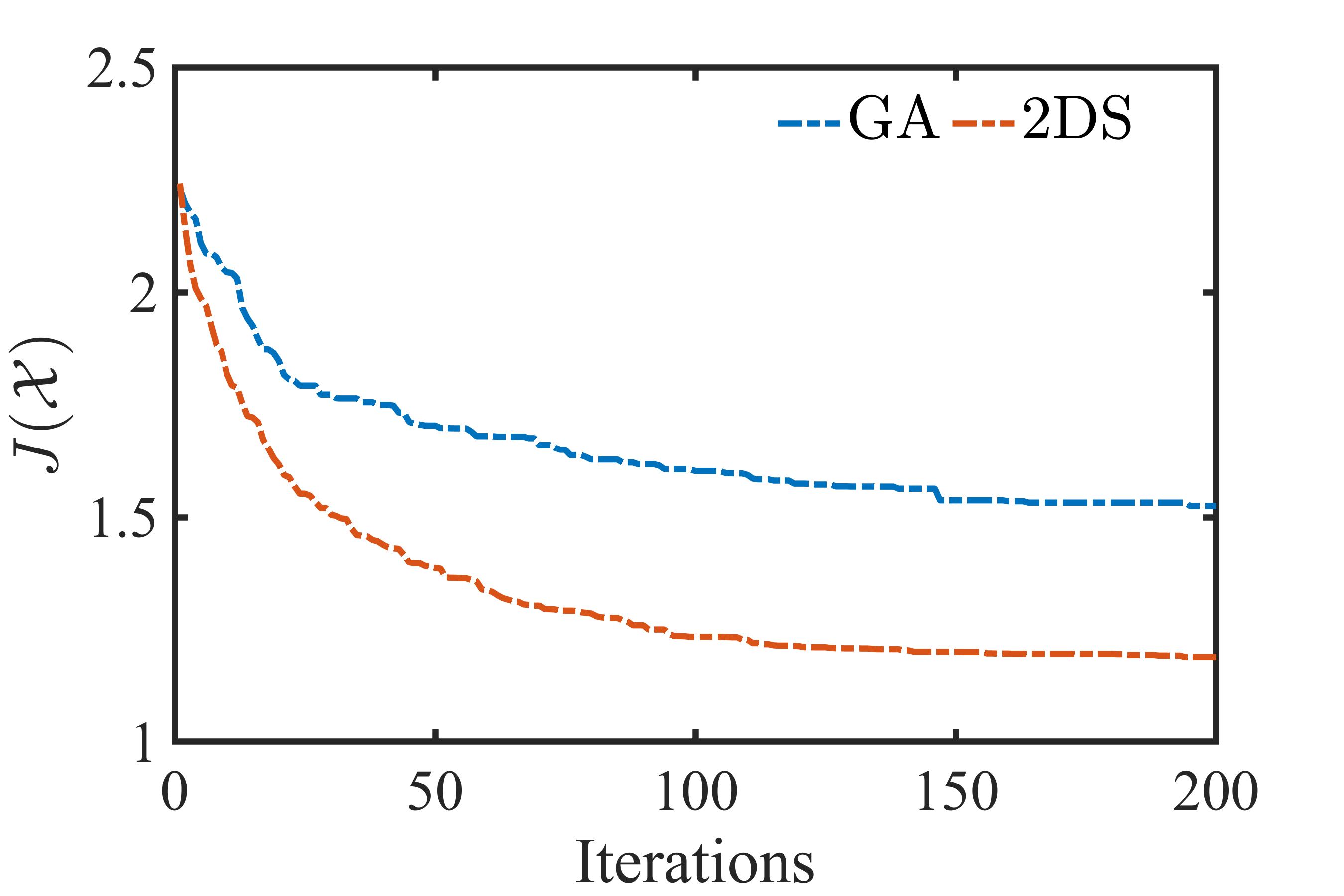}
  \caption{Criterion Function}
  \label{f:compJ}
\end{subfigure}
\hfill
\begin{subfigure}{.245\textwidth}
  \centering
  \includegraphics[width=\textwidth]{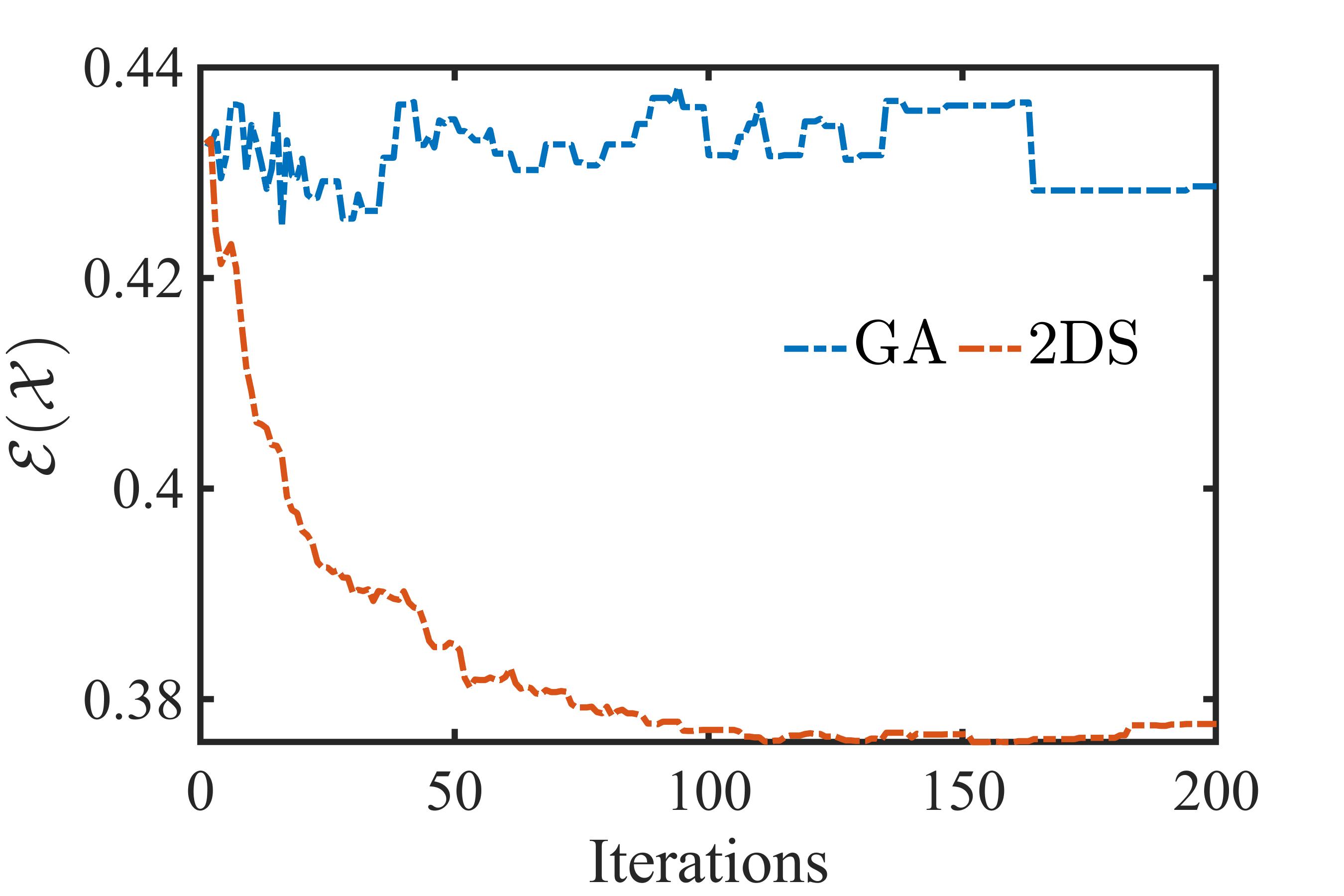}
  \caption{Error over $\mathcal{D}_{test}$}
  \label{f:compE}
\end{subfigure}
\hfill
\begin{subfigure}{.245\textwidth}
  \centering
  \includegraphics[width=\textwidth]{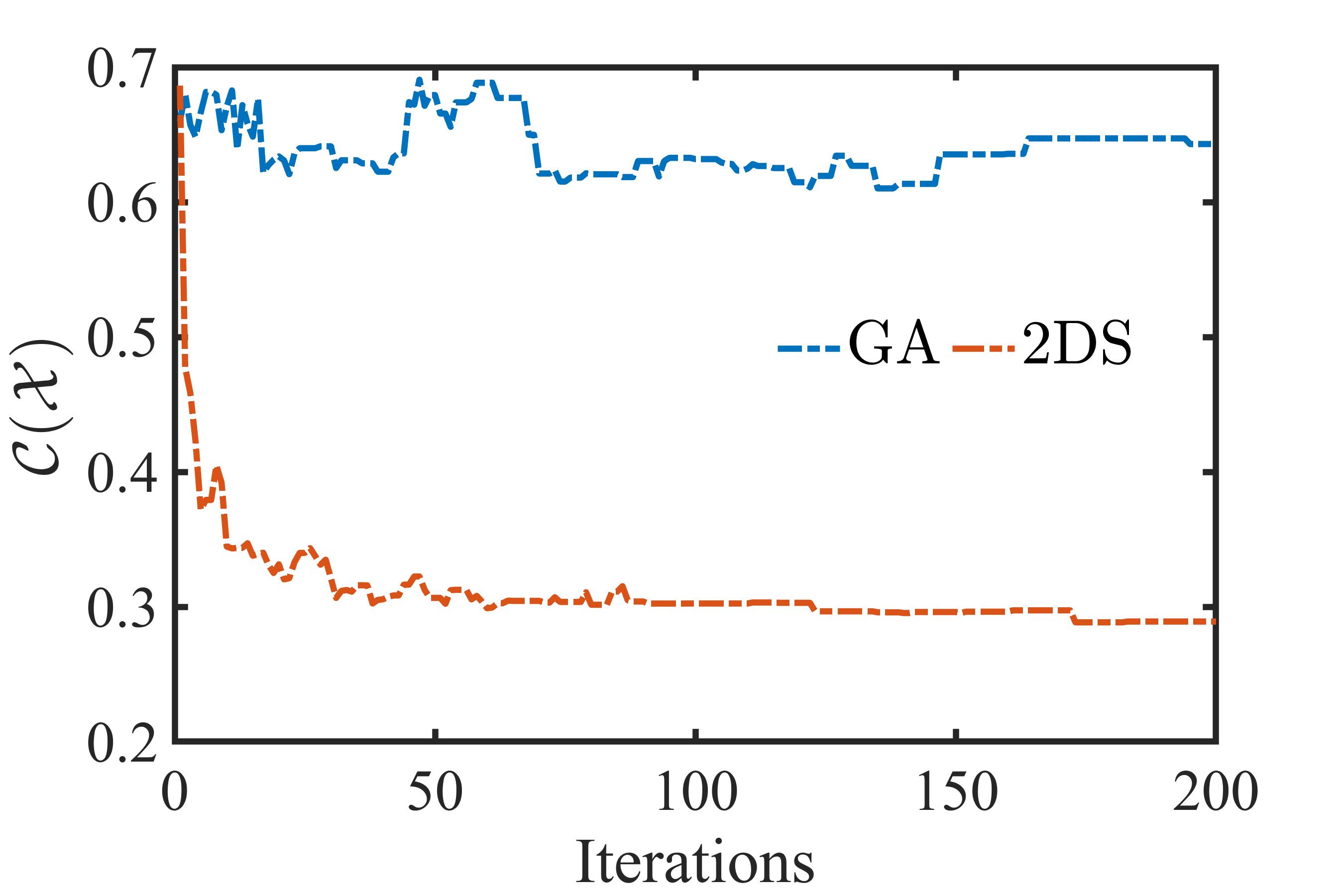}
  \caption{Complexity}
  \label{f:compC}
\end{subfigure}
\hfill
\begin{subfigure}{.245\textwidth}
  \centering
  \includegraphics[width=\textwidth]{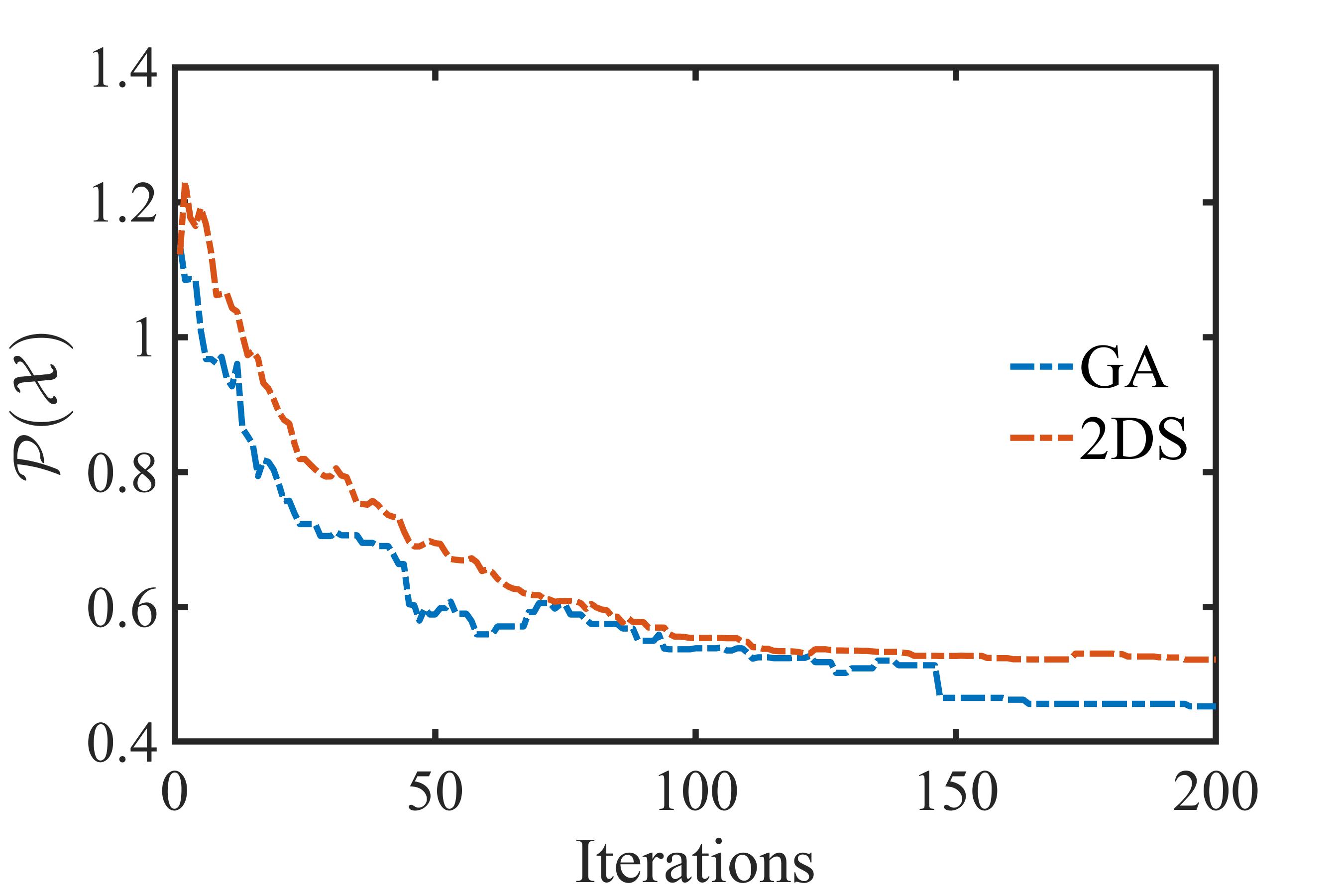}
  \caption{Penalty}
  \label{f:compP}
\end{subfigure}

\caption{Convergence plots averaged over $20$ runs of GA and 2DS under the \textit{mixed-learning} scenario, $\mathcal{LS}_\varepsilon$ and $\Theta_2$.}
\label{f:convGA+2DS}
\end{figure*}

To gain the further insight into dismal search performance of GA, its convergence analysis is studied and contrasted with that of 2DS. Fig.~\ref{f:convGA+2DS} show the convergence behaviors of the algorithm with the preference weight $\Theta_2$ and learning scenario $\mathcal{LS}_\varepsilon$. It is clear that GA could evolve neural architectures with slightly better penalty function, ($\mathcal{P}(\cdot)$) than 2DS (see Fig.~\ref{f:compP}); which indicates that GA could direct search better to meet the $\epsilon-$constraint requirements (see Section~\ref{subsec:multidataset_learning}). However, GA could not improve neural architectures in terms of either  \textit{efficacy} (see Fig.~\ref{f:compE}) or the complexity (see Fig.~\ref{f:compC}). 

It is interesting to note that the combination of mRmR feature selection and Fletcher Goss's rule could identify relatively less complex and more efficacious architecture than GA (see Table~\ref{t:benchmark}). To summarize, it is safe to infer that, GA could not evolve either efficacious or parsimonious neural architectures for the NASDAQ prediction problem being considered. These results further highlight the need for an effective search algorithm.

\begin{table*}[!t]
  \centering
  \caption{Best neural topology evolved out of 20 independent runs of GA with $\Theta_2$ under $\mathcal{LS}_\varepsilon$ learning scenario}
  \label{t:resGA}%
  \begin{adjustbox}{max width=\textwidth}
  \scriptsize
  \begin{threeparttable}
  
    \begin{tabular}{cccccccc}
    
    \toprule
    \multirow{2}[4]{*}{\makecell{\textbf{Selected}\\\textbf{Features,}\\ \boldmath{$\#X$}}} & \multicolumn{2}{c}{\textbf{Hidden Layer -1}} & \multicolumn{2}{c}{\textbf{Hidden Layer -2}} & \multirow{2}[4]{*}{\makecell{\textbf{Complexity,}\\\boldmath{$\mathcal{C}$}}} & \multirow{2}[4]{*}{\makecell{\textbf{Hold-out}\\\textbf{Accuracy}\\($\%$)}} & \multirow{2}[4]{*}{\makecell{\textbf{Hold-out}\\\textbf{MCC}}} \\[1ex]
    
    \cmidrule{2-5}  & \makecell{\textbf{Size,}\\\boldmath $s^1$} & \makecell{\textbf{Activation}\\\textbf{Function,} \boldmath$f^1$} & \makecell{\textbf{Size,}\\\boldmath $s^2$} & \makecell{\textbf{Activation}\\\textbf{Function,} \boldmath $f^2$} &       &       &  \\[1ex]
    \midrule
    
    27 & 107   & \textit{tansig} & 24    & \textit{logsig} & 0.6376 & 52.0  & 0.0381 \\
    \bottomrule
    \end{tabular}%

      
  \end{threeparttable}
 \end{adjustbox}
\end{table*}%
\subsection{Comparative Evaluation}
\label{subsec:comparative_evaluation}

In the final part of the investigation, the classification performance of the neural architectures identified using different approaches are compared. Given a lower acceptance threshold for accuracy, a \textit{naive-random} classifier is also included in comparison, which \textit{randomly} assigns \textit{up} or \textit{down} label for each test instance.  

This study considers non-parametric statistical tests, as the pre-requisites for the parametric tests such as multi-way ANOVA may not be satisfied by the search environment of this study~\citep{Derrac:Salvador:2011,Garcia:Salvador:2010}. In particular, Friedmann two-way analysis by ranks is first applied to test the \textit{null-hypothesis} that the classification performance of all the neural architectures do not differ. To this end, in each learning cycle (see Section~\ref{subsec:NeuEff}), the classification performance of each neural architecture over $\mathcal{D}_{hold}$ is ranked form \textit{best} ($1$) to \textit{worst} ($5$). Table~\ref{t:fried} gives the average rankings and $p-$\textit{value} of the Friedmann test for both accuracy and MCC. It is clear that \textit{null-hypothesis} is comfortably rejected. Further, the average rankings indicate the architecture identified by 2DS is the \textit{best} among all the compared architectures. Accordingly, 2DS architecture is being treated as the \textit{control} architecture in the following post-hoc analysis.

Next, a set of multiple \textit{null-hypotheses} are evaluated which stipulate the architecture being compared is significantly better than the architecture identified by 2DS. The Adjusted $p-$values (APV) to evaluate these \textit{null-hypotheses} are determined using the Hommel's post-hoc procedure~\citep{Derrac:Salvador:2011,Garcia:Salvador:2010}. The outcomes of this analysis are shown in Table~\ref{t:hommel}, which shows that all \textit{null-hypotheses} can safely be rejected for at $\alpha=0.05$ \textit{significance level}.
\begin{table}[!t]
  \centering
  \caption{Average Friedman Rankings of the Identified Architectures}
  \label{t:fried}%
    \scriptsize
    \begin{tabular}{ccccccc}
    \toprule
    \multirow{2}[4]{*}{\textbf{Results}} & \multicolumn{5}{c}{\textbf{Average Friedmann Rank}} & \multirow{2}{*}{\textbf{$p$-value}} \\
    \cmidrule{2-6}          & \textbf{Random} & \makecell{\textbf{Full} \boldmath$+$\\\textbf{FG's Rule}} & \makecell{\textbf{mRmR} \boldmath$+$\\\textbf{FG's Rule}} & \textbf{GA} & \textbf{2DS} &  \\[0.5ex]
    \midrule
    Accuracy & 3.65  & 3.40  & 2.93  & 3.63  & 1.40  & $1.08 \times 10^{-5}$ \\[0.5ex]
    MCC   & 3.65  & 3.58  & 2.83  & 3.55  & 1.40  & $7.22 \times 10^{-6}$\\
    \bottomrule
    \end{tabular}%
\end{table}%
\begin{table}[!t]
\centering
\scriptsize
\caption{Outcome of the Hommel's post-hoc procedure for $95\%$ confidence interval}
\label{t:hommel}

    \begin{tabular}{ccccc|cccc}
    \toprule
    \multirow{2}[4]{*}{\textbf{Topology}} & \multicolumn{4}{c}{\textbf{Accuracy}} & \multicolumn{4}{c}{\textbf{MCC}} \\
    \cmidrule{2-9}          & \makecell{\textbf{Test}\\\textbf{Static}} & \textbf{p-value} & \textbf{APV} & \boldmath{$H_0$} & \makecell{\textbf{Test}\\\textbf{Static}} & \textbf{p-value} & \textbf{APV} & \boldmath{$H_0$} \\[0.5ex]
    \midrule
    
    Random & 4.50  & 6.80E-06 & 0.013 & \xmark & 4.50  & 6.80E-06 & 0.013 & \xmark \\[1.8ex]
    \makecell{Full$+$\\FG's Rule}  & 4.00  & 6.33E-05 & 0.025 & \xmark & 4.35  & 1.36E-05 & 0.017 & \xmark \\[2.3ex]
    \makecell{mRmR $+$\\FG's Rule}  & 3.05  & 2.29E-03 & 0.050 & \xmark & 2.85  & 4.37E-03 & 0.050 & \xmark \\[2ex]
    GA    & 4.45  & 8.59E-06 & 0.017 & \xmark & 4.30  & 1.71E-05 & 0.025 & \xmark \\
 
    \bottomrule
    \end{tabular}%
\end{table}
\section{Conclusions}
\label{sec:conclusions}

The problem of day ahead prediction of NASDAQ index movement was explored from the neural design perspective. In particular, the attempts have been made to clarify the issues related to implications and possible remedies of disparate market behaviors prior to and during the ongoing COVID pandemic with respect to the neural architecture design. It is shown that any \textit{concordant} information underlying in the conflicting pre-COVID time window can be embedded in the neural design through $\epsilon-$constraint framework. Further, the problem of neural architecture design is extended and coupled with the issue feature selection. The ensuing \textit{extended} neural architecture search problem was formulated as the multi-criteria problem wherein the objective is to balance the architectural sparsity with the efficacy. A new search paradigm based on Two-Dimensional Swarms has been applied to solve the extended neural architecture search. The detailed comparative evaluation with GA and five empirical design rules convincingly demonstrate that 2DS can identify efficacious and parsimonious neural architectures.  

The investigation to study the inadvertent \textit{`data-poisoning'} effects associated with the pre-COVID time window highlights the need for a \textit{frequent} update of a predictive model as well as careful selection of the time window being considered for the supervised learning. The formulation of multi-objective \textit{transfer learning} or \textit{retraining} paradigm to reconcile evolving behavior of the market with its past tendencies can be explored for the further performance improvement.  

\appendix
\section{Empirical Rules for Neural Design}
\label{sec:app}

\begin{small}
\begin{itemize}
    \item Kolmogorov's Theorem : $[s^1, \ s^2] = \Big[(2 n_f + 1), \ 0\Big]$
    \smallskip
    \item Hush's Rule : $[s^1, \ s^2] = [c_1 \times n_f, \ c_2 \times m]$, where, $c_1 \in [2,4]$, $c_2 \in [2,3]$
    \smallskip
    \item Wang's Rule : $[s^1, \ s^2] = \Big[ \displaystyle\frac{2 \times n_f}{3}, \ 0\Big]$
    \smallskip
    \item Ripley's Rule : $[s^1, \ s^2] = \Big[ \displaystyle\frac{n_f + m}{2} , \ 0\Big]$
    \smallskip
    \item Fletcher-Goss's Rule : $[s^1, \ s^2] = \big[ 2\sqrt{n_f}+m, \ 0\big]$
    \smallskip
    \item Huang's Rule : $[s^1, \ s^2] = \Bigg[ \sqrt{(m+2)N} + \big(2\sqrt{\frac{N}{m+2}} \big), \ m\sqrt{\frac{N}{m+2}} \ \Bigg]$
\end{itemize}
\end{small}
\smallskip
where $n_f$, $m$, and $N$ respectively denote number of features, number of output classes/labels and total number of training samples.


%
%

\linespread{1}

\bibliographystyle{model5-names}
\biboptions{authoryear}

\end{document}